\PassOptionsToPackage{table}{xcolor}
\PassOptionsToPackage{pagebackref=false, breaklinks=true, letterpaper=true, colorlinks, bookmarks=false, citecolor=blue}{hyperref}
\documentclass[10pt,twocolumn,letterpaper]{article}

\usepackage{wacv}
\usepackage{times}

\usepackage{graphicx}
\usepackage{epstopdf}

\usepackage{amsmath}
\usepackage{amssymb}

\usepackage[style=ieee, backend=biber, citestyle=numeric-comp]{biblatex}
\usepackage{bm}

\usepackage{algorithm}
\usepackage{algorithmic}

\usepackage{mathtools}

\usepackage[subrefformat=parens,labelformat=parens]{subcaption}
\usepackage{booktabs}

\usepackage{multirow}
\usepackage{multicol}

\usepackage{dirtytalk}

\usepackage[table]{xcolor}
\makeatletter
\@namedef{ver@everyshi.sty}{}
\makeatother
\usepackage{tikz}
\usetikzlibrary{decorations.pathreplacing}
\usetikzlibrary{arrows.meta}

\usepackage{appendix}
\usepackage{tablefootnote}

\wacvfinalcopy % *** Uncomment this line for the final submission

\def\wacvPaperID{538} % *** Enter the wacv Paper ID here

\ifwacvfinal\fi
\begin{document}

    %%%%%%%%% TITLE
    \title{Uncertainty in Model-Agnostic Meta-Learning using Variational Inference}
    
    % Authors at the same institution
    %\author{First Author \hspace{2cm} Second Author \\
    % Authors at different institutions
    \author{Cuong Nguyen \\
        University of Adelaide\\
        {\tt\small cuong.nguyen@adelaide.edu.au}
        \and
        Thanh-Toan Do \\
        University of Liverpool\\
        {\tt\small thanh-toan.do@liverpool.ac.uk}
        \and
        Gustavo Carneiro \\
        University of Adelaide\\
        {\tt\small gustavo.carneiro@adelaide.edu.au}
    }
    
    \maketitle
    \ifwacvfinal\thispagestyle{empty}\fi
    
    \begin{abstract}
    We introduce a new, rigorously-formulated Bayesian meta-learning algorithm that learns a probability distribution of model parameter prior for few-shot learning. The proposed algorithm employs a gradient-based variational inference to infer the posterior of model parameters to a new task. Our algorithm can be applied to any model architecture and can be implemented in various machine learning paradigms, including regression and classification. We show that the models trained with our proposed meta-learning algorithm are well calibrated and accurate, with state-of-the-art calibration and classification results on two few-shot classification benchmarks (Omniglot, mini-ImageNet and tiered-ImageNet), and competitive results in a multi-modal task-distribution regression.
\end{abstract}
    
    \section{Introduction}
\label{sec:introduction}

Machine learning, in particular deep learning, has thrived during the last decade, producing results that were previously considered to be infeasible in several areas.  For instance, outstanding results have been achieved in speech and image understanding~\cite{hinton2012deep,graves2013speech,krizhevsky2012imagenet,simonyan2015very}, and medical image analysis~\cite{havaei2017brain}. However, the development of these machine learning methods typically requires a large number of training samples to achieve notable performance.  Such requirement contrasts with the human ability of quickly adapting to new learning tasks using few ``training" samples. This difference may be due to the fact that humans tend to exploit prior knowledge to facilitate the learning of new tasks, while machine learning algorithms often do not use any prior knowledge (e.g., training from scratch with random initialisation)~\cite{glorot2010understanding} or rely on weak prior knowledge to learn new tasks (e.g., training from pre-trained models)~\cite{rosenstein2005transfer}.  This challenge has motivated the design of machine learning methods that can make more effective use of prior knowledge to adapt to new learning tasks using few training samples~\cite{lake2015human}.

Such methods assume the existence of a latent distribution over classification or regression tasks that share a common structure.  This common structure means that solving many tasks can be helpful for solving a new task, sampled from the same task distribution, even if it contains a limited number of training samples.  For instance, in \textit{multi-task learning}~\cite{caruana1997multitask}, an agent simultaneously learns the shared representation of many related tasks and a main task that are assumed to come from the same domain. The extra information provided by this multi-task training tends to regularise the main task training, particularly when it contains few training samples. In \textit{domain adaptation} \cite{bridle1991recnorm,ben2010theory}, a learner transfers the shared knowledge of many training tasks drawn from one or several source domains to perform well on tasks (with small training sets) drawn from a target domain. \textit{Bayesian learning}~\cite{fei2006one} has also been explored, where prior knowledge is represented by a probability density function on the parameters of the visual classes' probability models.
In \textit{learning to learn} or \textit{meta-learning}~\cite{schmidhuber1987evolutionary,thrun1998learning}, a meta-learner extracts relevant knowledge from many tasks learned in the past to facilitate the learning of new future tasks.

From the methods above, meta-learning currently produces state-of-the-art results in many benchmark few-shot learning datasets~\cite{santoro2016meta,ravi2017optimization,munkhdalai2017meta,snell2017prototypical,finn2017model,yoon2018bayesian,zhang2018metagan,rusu2019meta}. Such success can be attributed to the way meta-learning leverages prior knowledge from several training tasks drawn from a latent distribution of tasks, where the objective is to perform well on unseen tasks drawn from the same distribution. However, a critical issue arises with the limited amount of training samples per task combined with the fact that most of these approaches~\cite{santoro2016meta,vinyals2016matching,ravi2017optimization,finn2017model,snell2017prototypical} do not try to estimate model uncertainty -- this may result in overfitting.  This issue has been recently addressed with Laplace approximation to estimate model uncertainty, involving the computationally hard estimation of a high-dimensional covariance matrix~\cite{grant2018recasting}, and with variational Bayesian learning~\cite{finn2018probabilistic,yoon2018bayesian} containing sub-optimal point estimate of model parameters and inefficient optimisation. 

In this work, we propose a new variational Bayesian learning by extending model-agnostic meta-learning (MAML)~\cite{finn2017model} based on a rigorous formulation that 
is efficient and does not require any point estimate of model parameters.  
In particular, compared to MAML~\cite{finn2017model}, our approach explores probability distributions over possible values of meta-parameters, rather than having a fixed value. Learning and prediction using our proposed method are, therefore, more robust due to the perturbation of learnt meta-parameters that coherently explains data variability.
%Unlike other Bayesian meta-learning methods, such as LLAMA~\cite{grant2018recasting}, PLATIPUS~\cite{finn2018probabilistic} and BMAML~\cite{yoon2018bayesian}, which required the computation of the Hessian~\cite{grant2018recasting}, or employed point estimate of model parameters that may reduce generalisation~\cite{finn2018probabilistic}, or used kernel matrix that complicates the computation and constrains the choice of variational function space~\cite{yoon2018bayesian}, our proposed method uses straightforward variational inference without the need of computing Hessian matrix, relying on point estimate or calculating a kernel matrix. Therefore, our method is simpler and computationally more efficient, while still generalising well.
Our evaluation shows that the models trained with our proposed meta-learning algorithm is at the same time well calibrated and accurate, with competitive results in terms of Expected Calibration Error (ECE) and Maximimum Calibration Error (MCE),
%robust with smaller predictive variances between tasks 
%\gustavocomment{I DON'T UNDERSTAND WHAT IS MEANT BY 'BETWEEN' TASKS} \cuongcomment{The reason is that we calculate the accuracy for each task, and report the avg of accuracies. Since we use model calibration, I drop this and emphasise the calibration}
while outperforming state-of-the-art methods in some few-shot classification benchmarks (Omniglot,  mini-ImageNet and tiered-ImageNet).
    \section{Related Work}
    Meta-learning has been studied for a few decades~\cite{schmidhuber1987evolutionary,naik1992meta,thrun1998learning}, and recently gained renewed attention with the use of deep learning methods. As meta-learning aims at the unique ability of learning how to learn, it has enabled the development of training methods with limited number of training samples, such as few-shot learning. Some notable meta-learning approaches include memory-augmented neural networks \cite{santoro2016meta}, deep metric learning \cite{vinyals2016matching,snell2017prototypical}, learn how to update model parameters~\cite{ravi2017optimization} and learn good prior using gradient descent update \cite{finn2017model}. %\gustavocomment{it is not clear what is meant by learned update} \cuongcomment{change from learned update to learned optimization algorithm} or gradient descent update~\cite{finn2017model}. 
    These approaches have generated some of the most successful meta-learning results, but they lack the ability to estimate model uncertainty. Consequently, their performances may suffer in uncertain environments and real world applications.
    
    % \subsection{Bayesian meta-learning}
    Bayesian meta-learning techniques have, therefore, been developed to incorporate uncertainty into model estimation. Among those, MAML-based meta-learning has attracted much of research interest due to the straightforward use of gradient-based optimisation of MAML. Grant et al.~\cite{grant2018recasting} use Laplace approximation to improve the robustness of MAML, but the need to estimate and invert the Hessian matrix makes this approach computationally challenging, particularly for large-scale models, such as the ones used by deep learning methods. Variational inference (VI) addresses such scalability issue -- remarkable examples of VI-based methods are PLATIPUS~\cite{finn2018probabilistic}, BMAML~\cite{yoon2018bayesian} and the methods similar to our proposal, Amortised meta-learner~\cite{ravi2018amortized} and VERSA~\cite{gordon2018metalearning}~\footnote{Amortised meta-learner~\cite{ravi2018amortized} and VERSA~\cite{gordon2018metalearning} have been developed in parallel to our proposed VAMPIRE.}. However, PLATIPUS optimises the lower bound of data prediction, leading to the need to approximate a joint distribution between the task-specific and meta parameters. This approximation complicates the implementation and requires a point estimate of the task-specific parameters to reduce the complexity of the estimation of this joint distribution. Employing point estimate may, however, reduce its ability to estimate  model uncertainty. BMAML uses a closed-form solution based on Stein Variational Gradient Descent (SVGD) that simplifies the task adaptation step, but it relies on the use of a kernel matrix, which increases its computational complexity. Amortised meta-learner applies variational approximation on both the meta-parameters and task-specific parameters, resulting in a challenging optimisation. VERSA takes a slightly different approach by employing an external neural network to learn the variational distribution for certain parameters, while keeping other parameters shared across all tasks. Another inference-based method is Neural Process~\cite{garnelo2018neural} that employs the train-ability of neural networks to model a Gaussian-Process-like distribution over functions to achieve uncertainty quantification in few-shot learning. However, due to the prominent weakness of Gaussian Process that suffers from cubic complexity to data size, this might limit the scalability of Neural Process and makes it infeasible for large-scale datasets.
    
    Our approach, in contrast, employs a straightforward variational approximation for the distribution of only the task-specific parameters, where we do not require the use of point estimate of any term, nor do we need to compute Hessian or kernel matrices or depend on an external network. Our proposed algorithm can be considered a rigorous and computationally efficient Bayesian meta-learning algorithm. A noteworthy non-meta-learning method that employs Bayesian methods is the neural statistician~\cite{edwards2017towards} that uses an extra variable to model data distribution within each task, and combines that information to solve few-shot learning problems. Our proposed algorithm, instead, does not introduce additional parameters, while still being able to extract relevant information from a small number of examples.

    \section{Methodology}
\label{sec:methodology}
    
    In this section, we first define and formulate the few-shot meta-learning problem. We then describe MAML, derive our proposed algorithm, and mention the similarities and differences between our method and recently proposed meta-learning methods that are relevant to our proposal.
    
    \subsection{Few-shot Learning Problem Setup}
    \label{sec:few_shot_problem}
    
        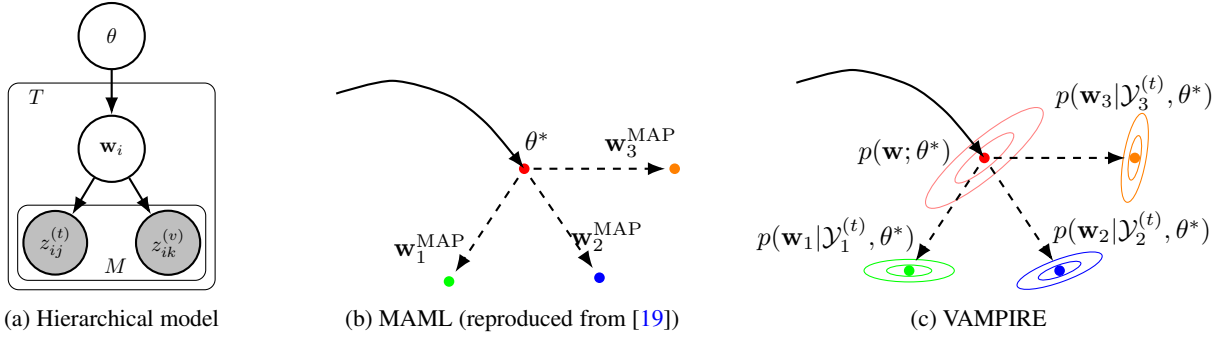
\begin{figure*}
            \centering
            \hfill
            \begin{subfigure}[t]{0.25\linewidth}
                \centering
                \begin{tikzpicture}[scale=0.5, every node/.style={scale=0.8}]
                   \node[circle, draw=black, thick, fill=gray!50] at (0, 0) (y_t) {\(z_{ij}^{(t)}\)};
                    \node[circle, draw=black, thick, fill=gray!50] at (3, 0) (y_v) {\(z_{ik}^{(v)}\)};
                    \draw[rounded corners] (-1, -1) rectangle (4, 1);
                    
                    \node[circle, draw=black, thick, minimum size=1.1cm] at (1.5, 2.5) (w_i) {\(\mathbf{w}_{i}\)};
            		
            		\draw[-Latex, thick] (w_i) -- (y_t);
            		\draw[-Latex, thick] (w_i) -- (y_v);
            		
        		    \draw[rounded corners] (-1.25, -1.25) rectangle (4.25, 4.25);
            		
            		\node[circle, draw=black, thick, minimum size=1.1cm] at (1.5, 5.5) (theta) {\(\theta\)};
            		\draw[-Latex, thick] (theta) -- (w_i);
            		
            		\node[label={[xshift=-1.25cm, yshift=0.5cm]\(T\)}] at (w_i) {};
            		\node[label={[xshift=1.cm, yshift=-0.75cm]\(M\)}] at (y_t) {};
        		\end{tikzpicture}
                \caption{Hierarchical model}
                \label{fig:hierarchical_model}
            \end{subfigure}
            \hfill
            \begin{subfigure}[t]{0.27\linewidth}
        	    \begin{tikzpicture}%[thick,scale=0.8, every node/.style={scale=0.8}]
        	        \node[label={[xshift=0.5em, yshift=0em]\(\theta^{*}\)}] (theta) at (3.5, 3){};
        	        \fill[color=red] (theta) circle [radius=2pt];
        	        \draw [thick, -Latex] plot [smooth] coordinates {(1, 4) (1.75, 4.17) (2.5, 3.97) (3, 3.6) (theta)};
        	        
        	        \node[label={[xshift=-0.25cm, yshift=0cm]\(\mathbf{w}_{1}^{\mathrm{MAP}}\)}] at (2.5, 1.5) (w1) {};
        	        \fill[color=green] (w1) circle [radius=2pt];
        	        
        	        \node[label={[xshift=0.1cm, yshift=0.1cm]\(\mathbf{w}_{2}^{\mathrm{MAP}}\)}] at (4.5, 1.55) (w2) {};
        	        \fill[color=blue] (w2) circle [radius=2pt];
        	        
        	        \node[label={[xshift=-0.45cm, yshift=-.1cm]\(\mathbf{w}_{3}^{\mathrm{MAP}}\)}] at (5.5, 3) (w3) {};
        	        \fill[color=orange] (w3) circle [radius=2pt];
        	        
        	        \draw[-Latex, dashed, thick] (theta) -- (w1);
        	        \draw[-Latex, dashed, thick] (theta) -- (w2);
        	        \draw[-Latex, dashed, thick] (theta) -- (w3);
        	    \end{tikzpicture}
        	   % \caption{}
        	    \caption{MAML (reproduced from \cite{finn2017model})}
                \label{fig:maml_initialization}
        	\end{subfigure}
        	\hfill
        	\begin{subfigure}[t]{0.35 \linewidth}
        	    \begin{tikzpicture}%[thick,scale=0.8, every node/.style={scale=0.8}]
        	        \node[label={[xshift=-3em, yshift=-1em]\(p(\mathbf{w}; \theta^{*})\)}] (theta) at (3.5, 3){};
        	        \fill[color=red] (theta) circle [radius=2pt];
        	        \foreach \i [count=\ni] in {1,...,2}
        	        {
        	           % \pgfmathtruncatemacro{\tmp}{100-\x*20}
        	            \draw[rotate around={40:(theta)}, draw=red!50] (theta) ellipse (\i*0.5 and \i*0.125);
        	        }
        	        \draw [thick, -Latex] plot [smooth] coordinates {(1, 4) (1.75, 4.17) (2.5, 3.97) (3, 3.6) (theta)};
        	        
        	        \node[label={[xshift=-2.75em, yshift=0cm]\(p(\mathbf{w}_{1} \vert \mathcal{Y}_{1}^{(t)}, \theta^{*})\)}] at (2.5, 1.5) (w1) {};
        	        \fill[color=green] (w1) circle [radius=2pt];
        	        \foreach \i [count=\ni] in {1,...,2}
        	        {
        	           % \pgfmathtruncatemacro{\tmp}{100-\x*20}
        	            \draw[rotate around={0:(w1)}, draw=green] (w1) ellipse (\i*0.3 and \i*0.075);
        	        }
        	        
        	        \node[label={[xshift=2.75em, yshift=0.25em]\(p(\mathbf{w}_{2} \vert \mathcal{Y}_{2}^{(t)}, \theta^{*})\)}] at (4.5, 1.5) (w2) {};
        	        \fill[color=blue] (w2) circle [radius=2pt];
        	        \foreach \i [count=\ni] in {1,...,2}
        	        {
        	           % \pgfmathtruncatemacro{\tmp}{100-\x*20}
        	            \draw[rotate around={20:(w3)}, draw=blue] (w2) ellipse (\i*0.3 and \i*0.075);
        	        }
        	        
        	        \node[label={[xshift=0em, yshift=1.1em]\(p(\mathbf{w}_{3} \vert \mathcal{Y}_{3}^{(t)}, \theta^{*})\)}] at (5.5, 3) (w3) {};
        	        \fill[color=orange] (w3) circle [radius=2pt];
        	        \foreach \i [count=\ni] in {1,...,2}
        	        {
        	           % \pgfmathtruncatemacro{\tmp}{100-\x*20}
        	            \draw[rotate around={80:(w3)}, draw=orange] (w3) ellipse (\i*0.3 and \i*0.075);
        	        }
        	        
        	        \draw[-Latex, dashed, thick] (theta) -- (w1);
        	        \draw[-Latex, dashed, thick] (theta) -- (w2);
        	        \draw[-Latex, dashed, thick] (theta) -- (w3);
        	    \end{tikzpicture}
        	   % \caption{}
        	    \caption{VAMPIRE}
        	    \label{fig:vampire_initialization}
        	\end{subfigure}
        	\caption{\protect\subref{fig:hierarchical_model} Hierarchical model of the few-shot meta-learning, aiming to learn \(\theta\) that parameterises prior \(p(\mathbf{w}_{i}; \theta)\), so that given a few data points \(z_{ij}^{(t)} = (x_{ij}^{(t)}, y_{ij}^{(t)})\) from the support set of task \(\mathcal{T}_{i}\), the model can quickly adapts and accurately predicts the output for the query set \(z_{ij}^{(v)} = (x_{ij}^{(v)}, y_{ik}^{(v)})\); \subref{fig:maml_initialization} and \subref{fig:vampire_initialization} Visualisation between MAML and VAMPIRE, respectively, where VAMPIRE extends the deterministic prior \(p(\mathbf{w}_{i}; \theta)\) and posterior \(p(\mathbf{w}_{i} \vert \mathcal{Y}_{i}^{(t)}, \theta)\) in MAML by using probabilistic distributions.}
        	\label{fig:hierarchical_model_and_visualisation}
        \end{figure*}
        
        While conventional machine learning paradigm is designed to optimise the performance on a single task, few-shot learning is trained on a set of conditional independent and identically distributed (i.i.d.) tasks given meta-parameters. The notation of \say{task environment} was formulated in \cite{baxter2000model}, where tasks are sampled from an unknown task distribution \(\mathcal{D}\) over a family of tasks.
        Each task \(\mathcal{T}_{i}\) in this family is indexed by \(i \in\{ 1,...,T \}\) and consists of a support set \(\{ \mathcal{X}_{i}^{(t)}, \mathcal{Y}_{i}^{(t)} \}\) and a query set \(\{ \mathcal{X}_{i}^{(v)}, \mathcal{Y}_{i}^{(v)} \}\), with \(\mathcal{X}_{i}^{(t)} = \{ \mathbf{x}_{ij}^{(t)} \}_{j=1}^M\) and \(\mathcal{Y}_{i}^{(t)} = \{ y_{ij}^{(t)} \}_{j=1}^M\) (\(\mathcal{X}_{i}^{(v)}\) and \(\mathcal{Y}_{i}^{(v)}\) are similarly defined). The aim of few-shot learning is to predict the output \(y_{ij}^{(v)}\) of the query input \(\mathbf{x}_{ij}^{(v)}\) given the small support set for task \(\mathcal{T}_{i}\) (e.g. \(M \le 20\)). We rely on a Bayesian hierarchical model~\cite{grant2018recasting} to model the few-shot meta-learning problem. In the graphical model shown in \figureautorefname{~\ref{fig:hierarchical_model}}, \(\theta\) denotes the meta parameters of interest, and \(\mathbf{w}_{i}\) represents the task-specific parameters for task \(\mathcal{T}_{i}\). One typical example of this modelling approach is MAML~\cite{finn2017model}, where \(\mathbf{w}_{i}\) are the neural network weights adapted to task \(\mathcal{T}_{i}\) by performing truncated gradient descent using the data from the support set \(\{\mathcal{X}_{i}^{(t)}, \mathcal{Y}_{i}^{(t)} \}\) and the initial weight values \(\theta\).
        
         The objective function of few-shot learning is, therefore, to find a meta-learner, parameterised by \(\theta\), across tasks sampled from \(\mathcal{D}\), as follows:
        \begin{equation}
            \theta^{*} = \arg \min_{\theta} \, - \frac{1}{T} \sum_{i=1}^{T} \ln p(\mathcal{Y}_{i}^{(v)} \vert \mathcal{Y}_{i}^{(t)}, \theta)
            \label{eq:few_shot_objective}
        \end{equation}
        where \(T\) denotes the number of tasks, and, hereafter, we simplify the notation by dropping the explicit dependence on \(\mathcal{X}_{i}^{(t)}\) and \(\mathcal{X}_{i}^{(v)}\) from the set of conditioning variables. Each term of the predictive probability on the right hand side of \eqref{eq:few_shot_objective} can be expanded by applying the sum rule of probability and lower-bounded by Jensen's inequality:
        \begin{equation*}
        \begin{aligned}[b]
            \ln p(\mathcal{Y}_{i}^{(v)} \vert \mathcal{Y}_{i}^{(t)}, \theta) & = \ln \mathbb{E}_{p(\mathbf{w}_{i} \vert \mathcal{Y}_{i}^{(t)}, \theta)} \left[ p(\mathcal{Y}_{i}^{(v)} \vert \mathbf{w}_{i}) \right] \ge \mathcal{L}_{i}^{(v)},
        \end{aligned}
        \label{eq:maml_likelihood}
        \end{equation*}
        where:
        \begin{equation}
            \mathcal{L}_{i}^{(v)}(\theta) = \mathbb{E}_{p(\mathbf{w}_{i} \vert \mathcal{Y}_{i}^{(t)}, \theta)} \left[ \ln p(\mathcal{Y}_{i}^{(v)} \vert \mathbf{w}_{i}) \right].
            \label{eq:task_specific_loss}
        \end{equation}
        
        Hence, instead of minimising the negative log-likelihood in~\eqref{eq:few_shot_objective}, we minimise the upper-bound of the corresponding negative log-likelihood which can be presented as:
        \begin{equation}
            \mathcal{L}^{(v)}(\theta) = -\frac{1}{T} \sum_{i=1}^{T} \mathcal{L}_{i}^{(v)}.
            \label{eq:log_likelihood}
        \end{equation}
        
        If each task-specific posterior, \(p(\mathbf{w}_{i} \vert \mathcal{Y}_{i}^{(t)}, \theta)\), is well-behaved, we can apply Monte Carlo to approximate the expectation in \eqref{eq:log_likelihood} by sampling model parameters \(\mathbf{w}_{i}\) from \(p(\mathbf{w}_{i} \vert \mathcal{Y}_{i}^{(t)}, \theta)\). Thus, depending on the formulation of the task-specific posterior \(p(\mathbf{w}_{i} \vert \mathcal{Y}_{i}^{(t)}, \theta)\), we can formulate different algorithms to solve the problem of few-shot learning. We review a deterministic method widely used in the literature in subsection~\ref{sec:maml}, and present our proposed approach in subsection~\ref{sec:vi}.
    
    \subsection{Point Estimate - MAML}
    \label{sec:maml}
        A simple way is to approximate \(p(\mathbf{w}_{i} \vert \mathcal{Y}_{i}^{(t)}, \theta)\) by a Dirac delta function at its local mode:
        \begin{equation}
            p(\mathbf{w}_{i} \vert \mathcal{Y}_{i}^{(t)}, \theta) = \delta(\mathbf{w}_{i} - \mathbf{w}_{i}^{\mathrm{MAP}}),
        	\label{eq:point_estimate_dirac_delta}
        \end{equation}
        where the local mode \(\mathbf{w}_{i}^{\mathrm{MAP}}\) can be obtained by using maximum a posterior (MAP):
        \begin{equation}
            \mathbf{w}_{i}^{\mathrm{MAP}} = \arg\max_{\mathbf{w}_{i}} \, \ln p(\mathcal{Y}^{(t)} \vert \mathbf{w}_{i}) + \ln p(\mathbf{w}_{i}; \theta).
        \end{equation}
        In the simplest case where the prior is also assumed to be a Dirac delta function: \(p(\mathbf{w}_{i}; \theta) = \delta(\mathrm{w}_{i} - \theta)\), and gradient descent is used, the local mode can be determined as:
        \begin{equation}
            \mathbf{w}_{i}^{\mathrm{MAP}} = \theta - \alpha \nabla_{\mathbf{w}_{i}} \left[-\ln p(\mathcal{Y}_{i}^{(t)} \vert \mathbf{w}_{i})\right],
            \label{eq:point_estimate_value}
        \end{equation}
        where $\alpha$ is the learning rate, and the truncated gradient descent consists of a single step of \eqref{eq:point_estimate_value} (the extension to a larger number of steps is trivial). Given the point estimate assumption in~\eqref{eq:point_estimate_dirac_delta}, the upper-bound of the negative log-likelihood in \eqref{eq:log_likelihood} can be simplified to:
        \begin{equation}
        \mathcal{L}^{(v)}(\theta) = \frac{1}{T} \sum_{i=1}^{T} -\ln p(\mathcal{Y}_{i}^{(v)} \vert \mathbf{w}_{i}^{\mathrm{MAP}}).
        \label{eq:maml_lower_bound}
        \end{equation}
        
        Minimising the upper-bound of the negative log-likelihood in \eqref{eq:maml_lower_bound} w.r.t. \(\theta\) represents the MAML algorithm~\cite{finn2017model}. This derivation also explains the intuition behind MAML, which finds a good initialisation of model parameters as illustrated in \figureautorefname{~\ref{fig:maml_initialization}}.
        
    \subsection{Gradient-based Variational Inference}
    \label{sec:vi}
        In contrast to the deterministic method presented in subsection~\ref{sec:maml}, we use a variational distribution \(q(\mathbf{w}_{i}; \lambda_{i})\), parameterized by \(\lambda_{i} = \lambda_{i}(\mathcal{Y}_{i}^{(t)}, \theta)\), to approximate the task-specific posterior \(p(\mathbf{w}_{i} \vert \mathcal{Y}_{i}^{(t)}, \theta)\). In variational inference, \(q(\mathbf{w}_{i}; \lambda_{i})\) can be obtained by minimising the following Kullback-Leibler (KL) divergence:
    	\begin{equation}
    	    \begin{aligned}[b]
    	        \lambda_{i}^* & = \arg \min_{\lambda_{i}} \mathrm{KL} \left[ q(\mathbf{w}_{i}; \lambda_{i}) \Vert p(\mathbf{w}_{i} \vert \mathcal{Y}_{i}^{(t)}, \theta) \right] \\
    	        & = \arg\min_{\lambda_{i}} \int q(\mathbf{w}_{i}; \lambda_{i}) \ln \frac{q(\mathbf{w}_{i}; \lambda_{i}) p(\mathcal{Y}_{i}^{(t)} \vert \theta) }{p(\mathcal{Y}_{i}^{(t)} \vert \mathbf{w}_{i}) p(\mathbf{w}_{i}; \theta)} \, d\mathbf{w}_{i} \\
    	        & = \arg \min_{\lambda_{i}} \mathcal{L}_{i}^{(t)} \left(\lambda_{i}, \theta \right) + \ln \underbrace{p(\mathcal{Y}^{(t)} \vert \theta)}_{\text{const. wrt } \lambda_{i}}.
    	    \end{aligned}
    	    \label{eq:kl_objective}
    	\end{equation}
        where:
        \begin{equation}
            \begin{aligned}[b]
                \mathcal{L}_{i}^{(t)} \left(\lambda_{i}, \theta \right) & = \mathrm{KL} \left[ q(\mathbf{w}_{i}; \lambda_{i}) \Vert p(\mathbf{w}_{i}; \theta) \right] \\
                & \quad + \mathbb{E}_{q(\mathbf{w}_{i}; \lambda_{i})} \left[ -\ln p(\mathcal{Y}_{i}^{(t)} \vert \mathbf{w}_{i}) \right].
            \end{aligned}
            \label{eq:variational_free_energy}
        \end{equation}
        
        The resulting cost function (excluding the constant term) \(\mathcal{L}_{i}^{(t)}\) is often known as the variational free energy (VFE). The first term of VFE can be considered as a regularisation that penalises the difference between the prior \(p(\mathbf{w}_{i}; \theta)\) and the approximated posterior \(q(\mathbf{w}_{i}; \lambda_{i})\), while the second term is referred as data-dependent part or likelihood cost. Exactly minimising the cost function in \eqref{eq:variational_free_energy} is computationally challenging, so gradient descent is used with \(\theta\) as the initialisation of \(\lambda_{i}\):
        \begin{equation}
            \lambda_{i} \gets \theta - \alpha \nabla_{\lambda_{i}} \mathcal{L}_{i}^{(t)} \left(\lambda_{i}, \theta \right),
            \label{eq:vi_gradient_update}
        \end{equation}
        where \(\alpha\) is the learning rate.
        
        Given the approximated posterior \(q(\mathbf{w}_{i}; \lambda_{i})\) with  parameter \(\lambda_{i}\) updated according to~\eqref{eq:vi_gradient_update}, we can calculate and optimise the upper-bound in~\eqref{eq:log_likelihood} to find a local-optimal meta-parameter \(\theta\).
        
        \begin{algorithm}[t]
        	\caption{VAMPIRE training}
        	\label{alg:vampire_train}
        	\begin{algorithmic}[1]
        		\REQUIRE task distribution \(\mathcal{D}\)
        		\REQUIRE Hyper-parameters: \(T, L_t, L_v, \alpha\) and \(\gamma\)
        		\STATE initialise \(\theta\)
        		\WHILE{\(\theta\) not converged}
        			\STATE sample a mini-batch of tasks \(\mathcal{T}_{i} \sim \mathcal{D}\), \(i=1:T\)
        			\FOR{each task \(\mathcal{T}_{i}\)}
        				\STATE \(\lambda_{i} \gets \theta\)
        				\STATE draw \(L_{t}\) samples \(\hat{\mathbf{w}}_{i}^{(l_{t})} \sim q(\mathbf{w}_{i}; \lambda_{i}), \, l_{t}=1:L_{t}\)
        				\STATE update: \(\lambda_{i} \gets \lambda_{i} - \frac{\alpha}{L_{t}} \nabla_{\lambda_{i}} \mathcal{L}_{i}^{(t)} \left(\lambda_{i}, \theta \right)\) \COMMENT{Eq~\eqref{eq:vi_gradient_update}} \label{alg_step:gradient_update}
        				\STATE draw \(L_{v}\) samples \(\hat{\mathbf{w}}_{i}^{(l_{v})} \sim q(\mathbf{w}_{i}; \lambda_{i}), \, l_{v} = 1:L_{v}\)
        				\STATE \(\mathcal{L}_{i}^{(v)} \left( \theta \right) = \frac{1}{L_v} \sum_{l_v=1}^{L_v} \ln p \left(\mathcal{Y}_{i}^{(v)} \vert \hat{\mathbf{w}}_{i}^{(l_v)} \right)\) \COMMENT{Eq.~\eqref{eq:task_specific_loss}} \label{alg_step:cost_function}
        			\ENDFOR
        			\STATE meta-update: \(\theta \gets \theta + \frac{\gamma}{T} \nabla_{\theta} \sum_{i=1}^{T} \mathcal{L}_{i}^{(v)} \left( \theta \right)\) \label{alg_step:meta_update}
        		\ENDWHILE
        	\end{algorithmic}
        \end{algorithm}
        
        In Bayesian statistics, the prior $p(\mathbf{w}_{i} | \theta)$ represents a modelling assumption, and the variational posterior $q(\mathbf{w}_{i}; \lambda_{i})$ is a flexible function that can be adjusted to achieve a good trade-off between performance and complexity. For simplicity, we assume that both $q(\mathbf{w}_{i}; \lambda_{i})$ and $p(\mathbf{w}_{i}; \theta)$ are Gaussian distributions with diagonal covariance matrices:
        \begin{equation}
            \begin{cases}
                p(\mathbf{w}_{i}; \theta) & = \mathcal{N}\left[\mathbf{w}_{i} \vert \bm{\mu}_{\theta}, \bm{\Sigma}_{\theta} = \mathrm{diag}(\bm{\sigma}_{\theta}^2)\right] \\
                q(\mathbf{w}_{i}; \lambda_{i}) & = \mathcal{N}\left[\mathbf{w}_{i} \vert \bm{\mu}_{\lambda_{i}}, \bm{\Sigma}_{\lambda_{i}} = \mathrm{diag}(\bm{\sigma}_{\lambda_{i}}^{2})\right],
            \end{cases}
            \label{eq:mle_gaussian_prior}
        \end{equation}
        where $\bm{\mu}_{\theta}, \bm{\mu}_{\lambda_{i}},  \bm{\sigma}_{\theta}, \bm{\sigma}_{\lambda_{i}} \in \mathbb{R}^d$, with \(d\) denoting the number of model parameters, and the operator $\mathrm{diag}(.)$ returns a diagonal matrix using the vector in the parameter.
        
        Given the prior \(p(\mathbf{w}_{i} | \theta)\) and the posterior \(q(\mathbf{w}_{i}; \lambda_{i})\) in~\eqref{eq:mle_gaussian_prior}, we can compute the KL divergence of VFE shown in \eqref{eq:variational_free_energy} by using either Monte Carlo sampling or a closed-form solution. According to \cite{blundell2015weight}, sampling model parameters from the approximated posterior \(q(\mathbf{w}_{i}; \lambda_{i})\) to compute the KL divergence term and optimise the cost function in \eqref{eq:variational_free_energy} does not perform better or worse than using the closed-form of the KL divergence between two Gaussian distributions. Therefore, we employ the closed-form formula of the KL divergence to speed up the training process.
        
        For numerical stability, we parameterise the standard deviation point-wisely as \({\sigma = \exp(\rho)}\) when performing gradient update for the standard deviations of model parameters. The meta-parameters \({\theta = (\bm{\mu}_{\theta}, \exp(\bm{\rho}_{\theta}))}\) are the initial mean and standard deviation of neural network weights, and the variational parameters \({\lambda_{i} = (\bm{\mu}_{\lambda_{i}}, \exp(\bm{\rho}_{\lambda_{i}}))}\) are the mean and standard deviation of those network weights optimised for task \(\mathcal{T}_{i}\). We also implement the re-parameterisation trick~\cite{kingma2014auto} when sampling the network weights from the approximated posterior to compute the expectation of the data log-likelihood in \eqref{eq:variational_free_energy}:
        \begin{equation}
            \mathbf{w}_{i} = \bm{\mu}_{\lambda_{i}} + \epsilon \odot \exp(\bm{\rho}_{\lambda_{i}}),
            \label{eq:reparameterization_trick}
        \end{equation}
        where \(\epsilon \sim \mathcal{N}(0, \mathbf{I}_{d})\), and \(\odot\) is the element-wise multiplication.  Given this direct dependency, the gradients of the cost function \(\mathcal{L}_{i}^{(t)}\) in \eqref{eq:variational_free_energy} with respect to \(\lambda_{i}\) can be derived as:
        \begin{equation}
            \begin{dcases}
            \nabla_{\mu_{\lambda_{i}}} \mathcal{L}_{i}^{(t)} & = \frac{\partial \mathcal{L}_{i}^{(t)}}{\partial \mathbf{w}_{i}} + \frac{\partial \mathcal{L}_{i}^{(t)}}{\partial \bm{\mu}_{\lambda_{i}}} \\
            \nabla_{\rho_{\lambda_{i}}} \mathcal{L}_{i}^{(t)} & = \frac{\partial \mathcal{L}_{i}^{(t)}}{\partial \mathbf{w}_{i}} \, \epsilon \odot \exp(\bm{\rho}_{\lambda_{i}}) + \frac{\partial \mathcal{L}_{i}^{(t)}}{\partial \bm{\rho}_{\lambda_{i}}}.
            \end{dcases}
        \end{equation}
        
        After obtaining the variational parameters \(\lambda_{i}\) in \eqref{eq:vi_gradient_update}, we can apply Monte Carlo approximation by sampling \(L_{v}\) sets of model parameters from the approximated posterior \(q(\mathbf{w}_{i}; \lambda_{i})\) to calculate and optimise the upper-bound in \eqref{eq:log_likelihood} w.r.t. \(\theta\). This approach leads to the general form of our proposed algorithm, named Variational Agnostic Modelling that Performs Inference for Robust Estimation (VAMPIRE), shown in Algorithm \ref{alg:vampire_train}.
    
    \subsection{Differentiating VAMPIRE and Other Bayesian Meta-learning Methods}
    \label{sec:differences}
        % VAMPIRE and MAML~\cite{finn2017model} are fundamentally different because MAML is a deterministic algorithm that employs point estimates on both the prior \(p(\mathbf{w}_{i}; \theta)\) and the task-specific posterior \(p(\mathbf{w}_{i} \vert \mathcal{Y}_{i}^{(t)}, \theta)\) (see \sectionautorefname{~\ref{sec:maml}}), while VAMPIRE  models these distributions in a probabilistically manner. This difference is depicted in \figureautorefname{~\ref{fig:maml_initialization}} and \figureautorefname{~\ref{fig:vampire_initialization}}.
        
        VAMPIRE is different from the \say{probabilistic MAML} - PLATIPUS \cite{finn2018probabilistic} in several ways. First, PLATIPUS uses VI to approximate the joint distribution \(p(\mathbf{w}_{i}, \theta \vert \mathcal{Y}_{i}^{(t)}, \mathcal{Y}_{i}^{(v)})\), while VAMPIRE uses VI to approximate the task-specific posterior \(p(\mathbf{w}_{i} \vert \mathcal{Y}_{i}^{(t)}, \theta)\). To handle the complexity of sampling from a joint distribution, PLATIPUS relies on the same point estimate of the task-specific posterior as MAML, as shown in~\eqref{eq:point_estimate_dirac_delta}. Second, to adapt to task \(\mathcal{T}_{i}\), PLATIPUS learns only the mean, without change the variance. In contrast, VAMPIRE learns both \(\bm{\mu}_{\theta}\) and \(\Sigma_{\theta}\) for each task \(\mathcal{T}_{i}\). Lastly, when adapting to a task, PLATIPUS requires 2 additional gradient update steps, corresponding to steps 7 and 10 of Algorithm 1 in~\cite{finn2018probabilistic}, while VAMPIRE needs only 1 gradient update step as shown in step \ref{alg_step:gradient_update} of Algorithm~\ref{alg:vampire_train}. Hence, VAMPIRE is based on a simpler formulation that does not rely on any point estimate, and it is also more flexible and efficient because it allows all meta-parameters to be learnt while performing less gradient-based steps.
        
        VAMPIRE is also different from the PAC-Bayes meta-learning method designed for multi-task learning~\cite{amit18meta} at the relation between the shared prior \(p(\mathbf{w}_{i}; \theta)\) and the variational task-specific posterior \(q(\mathbf{w}_{i}; \lambda_{i})\). While the PAC-Bayes meta-learning method does not relate the \say{posterior} to the \say{prior} as in the standard Bayesian analysis, VAMPIRE relates these two probabilities through a likelihood function by performing a fixed number of gradient updates as shown in \eqref{eq:vi_gradient_update}. Due to this discrepancy, the PAC-Bayes meta-learning needs to maintain all the task-specific posteriors, requiring more memory storage, consequently resulting in an un-scalable approach, especially when the number of tasks is very large. In contrast, VAMPIRE learns only the shared prior, and hence, is a more favourable method for large-scaled applications, such as few-shot learning.
        
        Our proposed algorithm is different from BMAML \cite{yoon2018bayesian} at the methods used to approximate task-specific posterior \(p(\mathbf{w}_{i} \vert \mathcal{Y}_{i}^{(t)}, \theta)\): BMAML is based on SVGD, while VAMPIRE is based on a variant of amortised inference. Although SVGD is a non-parametric approach that allows a flexible variational approximation, its downside is the computational complexity due to need to compute the kernel matrix, and high memory usage when increasing the number of particles. In contrast, our approach uses a straightforward variational method without any transformation of variables. One advantage of BMAML compared to our method in Algorithm~\ref{alg:vampire_train} is the use of Chaser Loss, which may be an effective way of preventing overfitting. Nevertheless, in principle, we can also implement the same loss for our proposed algorithm.
        
        VAMPIRE is different from Amortised Meta-learner~\cite{ravi2018amortized} at the data subset used to update the meta-parameters \(\theta\): whole data set of task \(\mathcal{T}_{i}\) in Amortised Meta-learner versus only the query subset \(\{ \mathcal{X}_{i}^{(v)}, \mathcal{Y}_{i}^{(v)} \}\) in VAMPIRE. This discrepancy is due to the differences in the objective function.  In particular, Amortised Meta-learner maximises the lower bound of marginal likelihood, while VAMPIRE maximises the predictive probability in \eqref{eq:few_shot_objective}. Moreover, when deriving a lower bound of marginal log-likelihood using VI~\cite[Derivation right before Eq. (1)]{ravi2018amortized}, the variational distribution \(q\) must be strictly greater than zero for all \(\theta\) and variational parameters. The assumption that approximates the variational distribution \(q(\theta; \psi)\) by a Dirac delta function made in Amortised ML~\cite[Eq. (4)]{ravi2018amortized} is, therefore, arguable.
        
        Another Bayesian meta-learning approach similar to VAMPIRE is VERSA~\cite{gordon2018metalearning}. The two methods are different at the methods modelling the parameters of interest \(\theta\). VAMPIRE relies on gradient update to relate the prior and posterior through likelihood function, while VERSA is based on an amortisation network to output the parameters of the variational distributions. To scale up to deep neural network models, VERSA models only the parameters of the last fully connected network, while leaving other parameters as point estimates that are shared across all tasks. As a result, VAMPIRE is more flexible since it does not need to define which parameters are shared or not shared, nor does it require any additional network.
        % Moreover, being model-agnostic, VAMPIRE is applicable to any machine learning model, while VERSA focuses only on deep neural networks.
    \section{Experimental Evaluation}
\label{sec:experiments}
    The goal of our experiments is to present empirical evaluation of VAMPIRE compared to state-of-art meta-learning approaches. Our experiments include both regression and few-shot classification problems. The experiments are carried out using the training procedure shown in Algorithm~\ref{alg:vampire_train}. All implementations of VAMPIRE use PyTorch~\cite{paszke2017automatic}.

    \subsection{Regression}
    \label{sec:regression}
        We evaluate VAMPIRE using a multi-modal task distribution where half of the data is generated from sinusoidal functions, while the other half is from linear functions~\cite{finn2018probabilistic}.
        % The amplitude and phase of the sinusoids are uniformly sampled from [0.1, 5] and [0, \(\pi\)], respectively, while the slope and intercept of the lines are sampled from [-3, 3]. Data is generated by uniformly sampling input from [-5, 5], and a zero-mean Gaussian noise with a standard deviation of 0.3 is added to the corresponding labels.
        A detailed configuration of the problem setup and the model used as well as additional visualisation results can be referred to \appendixname.
        
        % The model used is a 3-hidden fully connected neural network with 100 hidden units in each layer. The variational parameters \(\lambda_{i}\) is estimated by performing 5 gradient updates with learning rate \(\alpha = 0.001\). The meta-parameters \(\theta\) is obtained by using Adam \cite{kingma2015adam} with a fixed step size \(\gamma = 0.001\).
        
        \begin{figure}[t]
            \centering
            \includegraphics[width=0.9\linewidth]{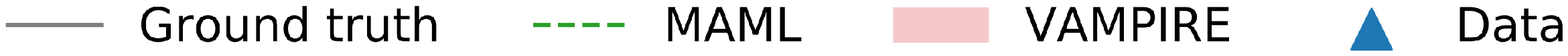}
            \hfill\\
            \vspace{-0.25em}
            \begin{subfigure}[t]{0.49\linewidth}
                \centering
                \includegraphics[width = \linewidth]{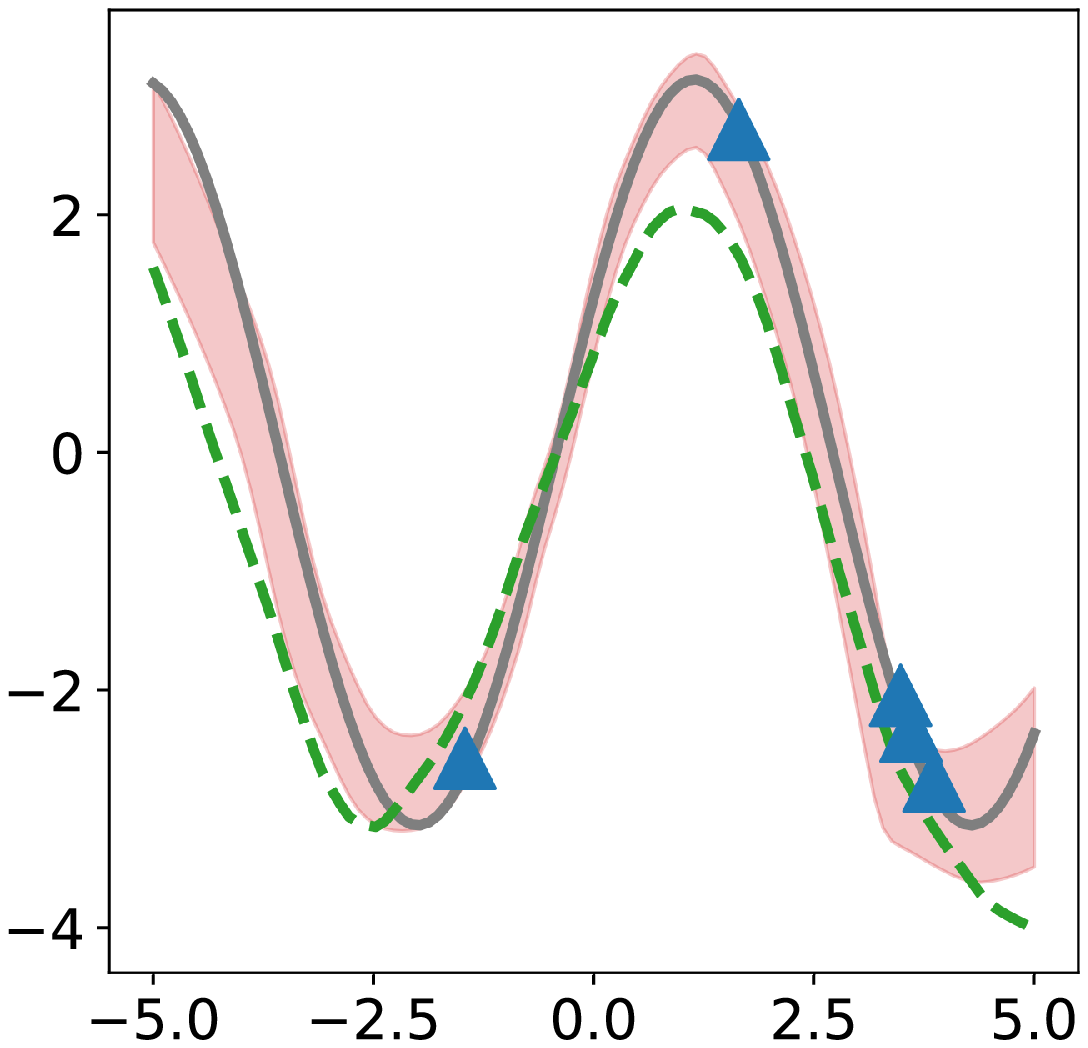}
                \caption{}
                \label{fig:regression_sine_1}
            \end{subfigure}
            \begin{subfigure}[t]{0.49\linewidth}
                \centering
                \includegraphics[width = \linewidth]{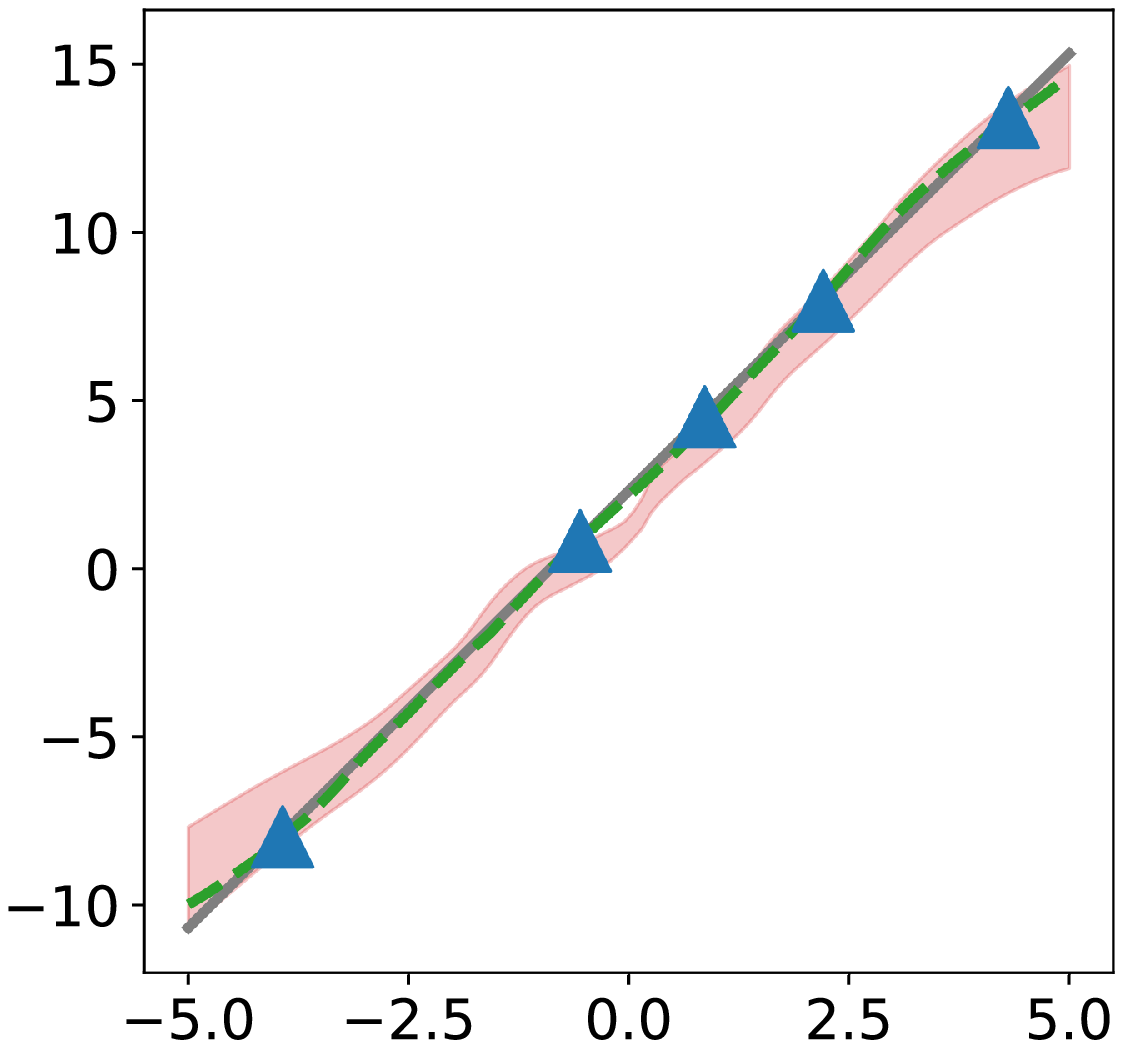}
                \caption{}
                \label{fig:regression_line_1}
            \end{subfigure}
            \hfill\\
            \begin{subfigure}[t]{0.49 \linewidth}
                \centering
                \includegraphics[width = \linewidth]{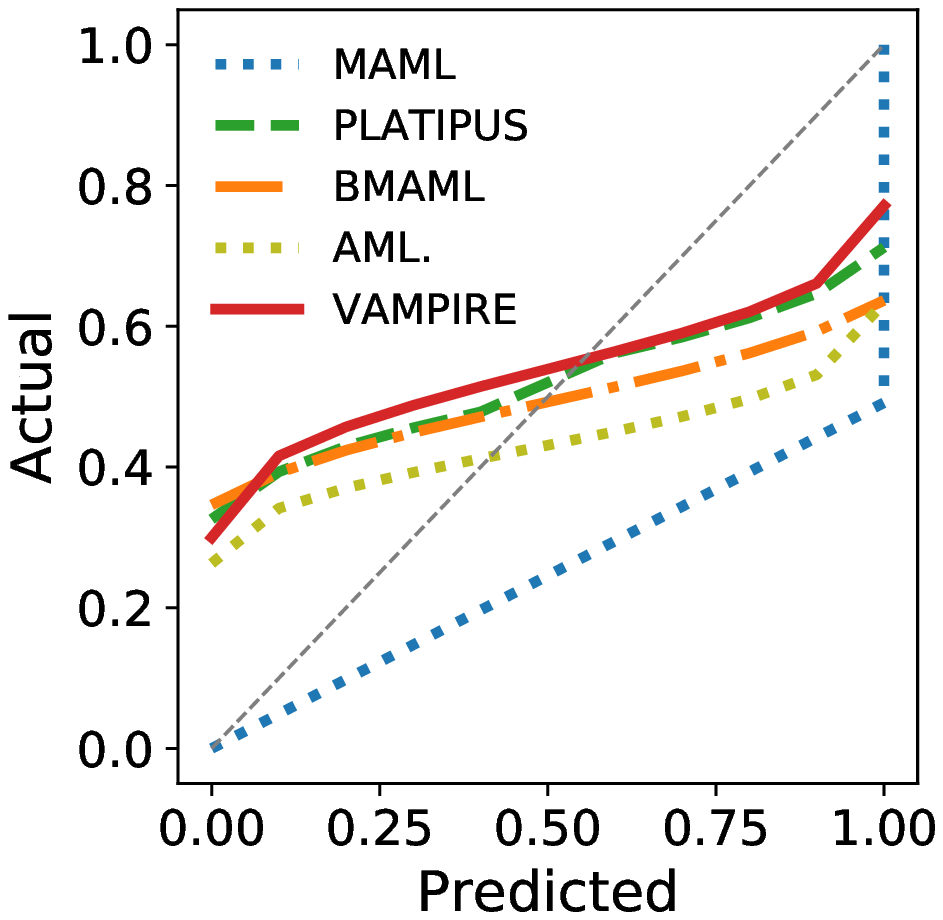}
                \caption{}
                \label{fig:regression_reliability}
            \end{subfigure}
            \begin{subfigure}[t]{0.49\linewidth}
                \includegraphics[width=\linewidth]{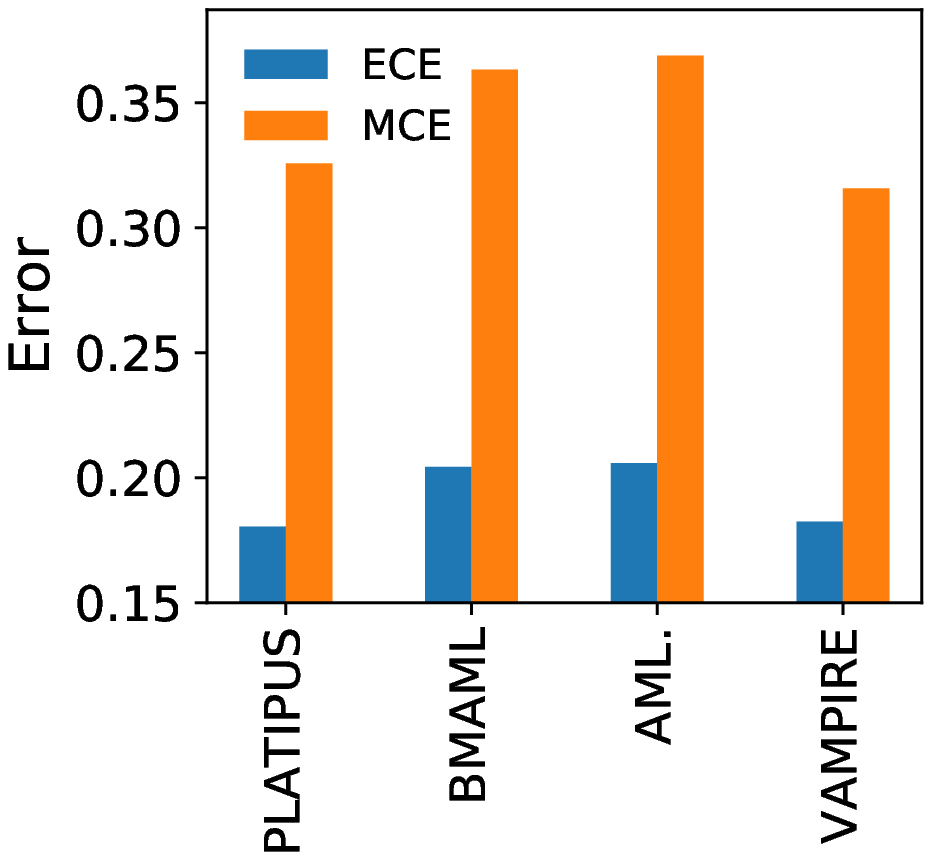}
                \caption{}
                \label{fig:regression_ece_mce}
            \end{subfigure}
            \caption{Qualitative and quantitative results on multi-modal data -- half of the tasks are generated from sinusoidal functions, and the other half are from linear functions: \protect \subref{fig:regression_sine_1} and \subref{fig:regression_line_1} visualisation of MAML and VAMPIRE, where the shaded area is the prediction made by VAMPIRE \(\pm\) 2\(\times\) standard deviation, \subref{fig:regression_reliability} reliability diagram of various meta-learning methods averaged over 1000 tasks, and \subref{fig:regression_ece_mce} ECE and MCE of the Bayesian meta-learning methods.}
            \label{fig:regression_sine_line}
        \end{figure}
        
        The results in \figureautorefname{s~\ref{fig:regression_sine_1}} and \ref{fig:regression_line_1} show that VAMPIRE can effectively reason which underlying function generates the training data points as the predictions are all sinusoidal or linear. In addition, VAMPIRE is able to vary the prediction variance, especially when there is more uncertainty in the training data. In contrast, due to the deterministic nature, MAML can only output a single value at each input.
        
        To quantitatively compare the performance between VAMPIRE and other few-shot meta-learning methods, we use the reliability diagram which is based on the quantile calibration for regression~\cite{song19distribution}. A model is perfectly calibrated when its predicted probability equals to the actual probability, resulting in a curve that is well-aligned with the diagonal \(y = x\). We re-implement some few-shot meta-learning methods, train until convergence, and plot their reliability diagram for 1000 tasks in \figureautorefname{~\ref{fig:regression_reliability}}. To have a fair comparison, BMAML is trained without Chaser Loss, and Amortised Meta-learner is trained with a uniform hyper-posterior. Due to the deterministic nature, MAML is presented as a single point connecting with the two extreme points. For a further quantitative comparison, we also plot the expected calibration error (ECE), which averages the absolute errors measuring from the diagonal, and the maximum calibration error (MCE), which returns the maximum of absolute errors in \figureautorefname{~\ref{fig:regression_ece_mce}}.
        Overall, in terms of ECE and MCE, the model trained with VAMPIRE is better than BMAML and Amortised Meta-learner, while competitive with PLATIPUS. The performance of BMAML could be higher if more particles and Chaser Loss are used. Another observation is that Amortised Meta-learner has slightly lower performance than MAML, although the training procedures of the two methods are very similar. We hypothesise that this is due to overfitting induced by using the whole training data subset that includes \(\{\mathcal{X}_{i}^{(t)}, \mathcal{Y}_{i}^{(t)}\}\), while MAML and VAMPIRE use only the query data subset \(\{\mathcal{X}_{i}^{(v)}, \mathcal{Y}_{i}^{(v)}\}\) to train the meta-parameters, which is consistent between the training and testing scenarios.

    \subsection{Few-shot Classification}
    \label{sec:classification}
        The experiments in this sub-section are based on the \(N\)-way \(k\)-shot learning task, where a meta learner is trained on many related tasks containing \(N\) classes and small training sets of \(k\) samples for each class (i.e., this is the size of \(\mathcal{Y}_{i}^{(t)}\)). We benchmark our results against the state of the art on the data sets Omniglot \cite{lake2015human}, mini-ImageNet \cite{vinyals2016matching,ravi2017optimization} and tiered-ImageNet~\cite{ren2018meta}.
        
        \begin{table*}[t!]
            \begin{center}
                \begin{small}
                    \begin{sc}
                        \begin{tabular}{l c c c c}
                            \toprule
                            \multirow{2}{*}{} & \multicolumn{2}{c}{\textbf{5-way}} & \multicolumn{2}{c}{\textbf{20-way}} \\
                             & 1-shot & 5-shot & 1-shot & 5-shot \\
                            \midrule
                            \midrule
                            \multicolumn{5}{l}{\textbf{Omniglot} \cite{lake2015human} - \textbf{Original Split, standard 4-layer CNN}}\\
                            \midrule
                            MAML & \(\mathbf{96.68 \pm 0.57}\) & \(98.33 \pm 0.22\) & \(84.38 \pm 0.64\) & \(\mathbf{96.32 \pm 0.17}\) \\
                            \rowcolor{gray!30} \textbf{VAMPIRE}  & \(96.27 \pm 0.38\) & \(\mathbf{98.77 \pm 0.27}\) & \(\mathbf{86.60 \pm 0.24}\) & \(96.14 \pm 0.10\) \\
                            \midrule
                            \midrule
                            \multicolumn{5}{l}{\textbf{Omniglot} \cite{lake2015human} - \textbf{Random Split, standard 4-layer CNN}}\\
                            \midrule
                            Matching nets \cite{vinyals2016matching} & \(98.1\) & \(98.9\) & \(93.8\) & \(98.5\) \\
                            Proto. nets \cite{snell2017prototypical}~\footnotemark[2] & \(98.8\) & \(99.7\) & \(96.0\) & \(\mathbf{98.9}\) \\
                            MAML \cite{finn2017model} & \(\mathbf{98.7 \pm 0.4}\) & \(\mathbf{99.9 \pm 0.1}\) & \(\mathbf{95.8 \pm 0.3}\) & \(\mathbf{98.9 \pm 0.2}\) \\
                            \rowcolor{gray!30} \textbf{VAMPIRE}  & \(98.43 \pm 0.19\) & \(99.56 \pm 0.08\) & \(93.20 \pm 0.28\) & \(98.52 \pm 0.13\) \\
                            \midrule
                            \midrule
                            \multicolumn{5}{l}{\textbf{Omniglot} \cite{lake2015human} - \textbf{Random Split, non-standard CNNs}}\\
                            \midrule
                            Siamese nets \cite{koch2015siamese} & \(97.3\) & \(98.4\) & \(88.2\) & \(97.0\) \\
                            Neural statistician \cite{edwards2017towards} & \(98.1\) & \(99.5\) & \(93.2\) & \(98.1\) \\
                            Memory module \cite{kaiser2017learning} & \(98.4\) & \(99.6\) & \(95.0\) & \(98.6\) \\
                            Relation nets \cite{Sung_2018_CVPR} & \(99.6 \pm 0.2\) & \(\mathbf{99.8 \pm 0.1}\) & \(97.6 \pm 0.2\) & \(\mathbf{99.1 \pm 0.1}\) \\
                            VERSA \cite{gordon2018metalearning} & \(\mathbf{99.70 \pm 0.20}\) & \(99.75 \pm 0.13\) & \(\mathbf{97.66 \pm 0.29}\) & \(98.77 \pm 0.18\) \\
                            \bottomrule
                        \end{tabular}
                    \end{sc}
                \end{small}
            \end{center}
            \vspace{-0.5em}
            \caption{Few-shot classification accuracy (in percentage) on Omniglot, tested on 1000 tasks and reported with 95\% confidence intervals. The results of VAMPIRE are competitive to the state-of-the-art baselines which are carried out on a standard 4-convolution-layer neural networks. The top of the table contains 
            methods trained on the original split defined in~\cite{lake2015human}, while the middle part contains methods using a standard 4-layer CNN trained on random train-test split. The bottom part presents results of different methods using different network architectures, or requiring external modules and additional parameters trained on random split. Note that the Omniglot results on random split cannot be fairly compared.}
            \label{tab:omniglot_results}
            % \vspace{-1em}
        \end{table*}
        
        Omniglot contains 1623 different handwritten characters from 50 different alphabets, where each one of the characters was drawn online via Amazon's Mechanical Turk by 20 different people \cite{lake2015human}. Omniglot is often split by randomly picking 1200 characters for training and the remaining for testing~\cite{ravi2017optimization,finn2017model,snell2017prototypical}. However, for language character classification, this random split may be unfair since knowing a character of an alphabet may facilitate the learning of other characters in the same alphabet. The original train-test split defined in~\cite{lake2015human} suggests 30 alphabets for training and 20 alphabets for testing -- such split clearly avoids potential information leakage from the training set to the testing set. We run experiments using both splits to compare with state-of-the-art methods and to perform testing without any potential data leakage. As standardly done in the literature, our training includes a data augmentation based on rotating the samples by multiples of 90 degrees, as proposed in~\cite{santoro2016meta}. Before performing experiments, all Omniglot images are down-sampled to 28-by-28 pixels to be consistent with the reported works in the meta-learning literature~\cite{vinyals2016matching,ravi2017optimization,finn2017model}.
        
        Mini-ImageNet was proposed in~\cite{vinyals2016matching} as an evaluation for few-shot learning. It consists of 100 different classes, each having 600 colour images taken from the original ImageNet data set~\cite{imagenet_cvpr09}. We use the train-test split reported in~\cite{ravi2017optimization} that consists of 64 classes for training, 16 for validation, and 20 for testing. Similarly to  Omniglot, the examples in mini-ImageNet are pre-processed by down-sampling the images to 84-by-84 pixels to be consistent with previous works in the literature.
        
        Tiered-ImageNet~\cite{ren2018meta} is a larger subset of ImageNet that has 608 classes grouped into 34 high-level categories. We use the standard train-test split that consists of 20, 6, and 8 categories for training, validation and testing. The experiments on tiered-ImageNet is carried with input as features extracted by a residual network that was pre-trained on data and classes from training meta-set \cite[Section 4.2.2]{rusu2019meta}.
        
        For Omniglot and mini-ImageNet, we use the same network architecture of state-of-the-art methods~\cite{vinyals2016matching,ravi2017optimization,finn2017model}. The network consists of 4 hidden convolution modules, each containing 64 3-by-3 filters, followed by batch normalisation \cite{ioffe2015batch}, ReLU activation, and a 2-by-2 strided convolution. For the mini-ImageNet, the strided convolution is replaced by a 2-by-2 max-pooling layer, and only 32 filters are used on each convolution layer to avoid over-fitting~\cite{ravi2017optimization,finn2017model}. For tiered-ImageNet, we use a 2-hidden-layer fully-connected network with 128 and 32 hidden units. Please refer to \appendixname for detailed description on the configuration and the hyperparameters used.
        
        \footnotetext[2]{Trained with 60-way episodes.}
        
        The \(N\)-way \(k\)-shot classification accuracy measured on Omniglot, and mini-ImageNet, tiered-ImageNet data sets are shown in \tableautorefname{s~\ref{tab:omniglot_results}} and \ref{tab:miniImageNet_results}, respectively.
        % Although our goal is to compare algorithms by using the standard 4-layer CNN, we also include the state-of-the-art methods on mini-ImageNet that employ much deeper networks with various architectures to provide a broad view of the few-shot image classification in \appendixname{~\ref{apdx:classification}}. 
        Overall, the results of VAMPIRE are competitive to the state-of-the-art methods that use the same network architecture~\cite{vinyals2016matching,ravi2017optimization,finn2017model}.
        
        On Omniglot, our results on a random train-test split are competitive in most scenarios. Our proposed method outperforms some previous works in few-shot learning, such as siamese networks \cite{koch2015siamese}, matching networks \cite{vinyals2016matching} and memory models \cite{kaiser2017learning}, although they are designed with a focus on few-shot classification. Our result on the 20-way 1-shot is slightly lower than prototypical networks~\cite{snell2017prototypical} and VERSA~\cite{gordon2018metalearning}, but prototypical networks need more classes (higher \say{way}) per training episode to obtain advantageous results and VERSA requires an additional amortised networks to learn the variational distributions. Our results are also slightly lower than MAML, potentially due to the difference of train-test split. To obtain a fair comparison, we run the public code provided by MAML's authors, and measure its accuracy on the original split suggested in~\cite{lake2015human}. Using this split, VAMPIRE achieves competitive performance, and outperforms MAML in some cases.
        
        On mini-ImageNet, VAMPIRE outperforms all reported methods that use the standard 4-layer CNN architecture on the 1-shot tests, while being competitive on the 5-shot episodes. Prototypical Networks achieve a higher accuracy on the 5-shot tests due to, again, the use of extra classes during training. Although our work does not aim to achieve the state-of-the-art results in few-shot learning, we also run an experiment using as input features extracted by a residual network that was pre-trained on data and classes from training meta-set \cite[Sect. 4.2.2]{rusu2019meta}, and present the results, including the state-of-the-art methods that employ much deeper networks with various architectures, in \appendixname. Note that deeper networks tend to reduce intra-class variation, resulting in a smaller gap of performance among many meta-learning methods~\cite{chen2018a}.
        % Compared to some recent methods such as TADAM \cite{oreshkin2018tadam} and LEO \cite{rusu2019meta}, our results achieve the current state-of-the-art results in 1-shot, and are competitive in 5-shot. Note that the results in the bottom part of \tableautorefname{~\ref{tab:miniImageNet_results}} are not directly comparable since the network architecture used and the training setting vary in each method.
        
        On tiered-ImageNet, VAMPIRE outperforms many methods published previously by a large margin on both 1- and 5-shot settings.
        
        \begin{table}[t]
            \begin{center}
                \begin{small}
                    \begin{sc}
                        \begin{tabular}{p{10em} c c}
                            \toprule
                            \multirow{2}{*}{} & \multicolumn{2}{c}{\textbf{Mini-ImageNet}~\cite{ravi2017optimization}} \\
                             & 1-shot & 5-shot\\
                            \midrule
                            \midrule
                            \multicolumn{3}{l}{\textbf{Standard 4-block CNN} }\\
                            \midrule
                            Matching nets \cite{vinyals2016matching} & \(43.56 \pm 0.84\) & \(55.31 \pm 0.73\) \\
                            Meta-learner LSTM \cite{ravi2017optimization} & \(43.44 \pm 0.77\) & \(60.60 \pm 0.71\) \\
                            MAML \cite{finn2017model} & \(48.70 \pm 1.84\) & \(63.15 \pm 0.91\) \\
                            Proto. nets \cite{snell2017prototypical} \tablefootnote{Trained with 30-way episodes for 1-shot classification and 20-way episodes for 5-shot classification} & \(49.42 \pm 0.78\) & \(\mathbf{68.20 \pm 0.66}\) \\
                            LLAMA \cite{grant2018recasting} & \(49.40 \pm 1.83\) & \_ \\
                            PLATIPUS \cite{finn2018probabilistic} & \(50.13 \pm 1.86\) & \_ \\
                            BMAML~\cite{yoon2018bayesian}\tablefootnote{Produced locally} & \(49.17 \pm 0.87\) & \(64.23 \pm 0.69\) \\
                            Amortised ML \cite{ravi2018amortized} & \(45.00 \pm 0.60\) & \_ \\
                            \rowcolor{gray!30}\textbf{VAMPIRE} & \(\mathbf{51.54 \pm 0.74}\) & \(64.31 \pm 0.74\) \\
                            \bottomrule
                            \toprule
                            \multirow{2}{*}{} & \multicolumn{2}{c}{\textbf{Tiered-ImageNet} \cite{ren2018meta}}\\
                             & 1-shot & 5-shot \\
                            \midrule
                            MAML \cite{liu2018transductive} & \(51.67 \pm 1.81\) & \(70.30 \pm 0.08\) \\
                            Proto. Nets \cite{ren2018meta} & \(53.31 \pm 0.89\) & \(72.69 \pm 0.74\) \\
                            Relation Net \cite{liu2018transductive} & \(54.48 \pm 0.93\) & \(71.32 \pm 0.78\) \\
                            Trns. Prp. Nets \cite{liu2018transductive} & \(57.41 \pm 0.94\) & \(71.55 \pm 0.74\) \\
                            LEO \cite{rusu2019meta} & \(66.33 \pm 0.05\) & \(81.44 \pm 0.09\) \\
                            MetaOptNet \cite{lee2019meta} & \(65.81 \pm 0.74\) & \(81.75 \pm 0.53\) \\
                            \rowcolor{gray!30}\textbf{VAMPIRE} & \(\mathbf{69.87 \pm 0.29}\) & \(\mathbf{82.70 \pm 0.21}\) \\
                            \bottomrule
                        \end{tabular}
                    \end{sc}
                \end{small}
            \end{center}
            \vspace{-0.75em}
            \caption{The few-shot 5-way classification accuracy results (in percentage) of VAMPIRE averaged over 600 mini-ImageNet tasks and 5000 tiered-ImageNet tasks are competitive to the state-of-the-art methods.}
            \label{tab:miniImageNet_results}
        \end{table}
        
        % \footnotetext[3]{Trained with 30-way episodes for 1-shot classification and 20-way episodes for 5-shot classification}
        
\begin{figure}[t]
            \centering
            % \begin{subfigure}{0.495\linewidth}
            %     \includegraphics[width=\linewidth]{img/reliability_maml.eps}
            %     \caption{MAML}
            %     \label{fig:reliability_diagram_maml}
            % \end{subfigure}
            % \hfill
            % \begin{subfigure}{0.495\linewidth}
            %     \includegraphics[width=\linewidth]{img/reliability_platipus.eps}
            %     \caption{PLATIPUS}
            %     \label{fig:reliability_diagram_platipus}
            % \end{subfigure}
            % \\
            % \begin{subfigure}{0.495\linewidth}
            %     \includegraphics[width=\linewidth]{img/reliability_aml.eps}
            %     \caption{Amortised Meta-learner}
            %     \label{fig:reliability_diagram_aml}
            % \end{subfigure}
            % \hfill
            % \begin{subfigure}{0.495\linewidth}
            %     \includegraphics[width=\linewidth]{img/reliability_vampire.eps}
            %     \caption{VAMPIRE}
            %     \label{fig:reliability_diagram_vampire}
            % \end{subfigure}
            % \\
            \begin{subfigure}{0.485\linewidth}
                \centering
                \includegraphics[width = \linewidth]{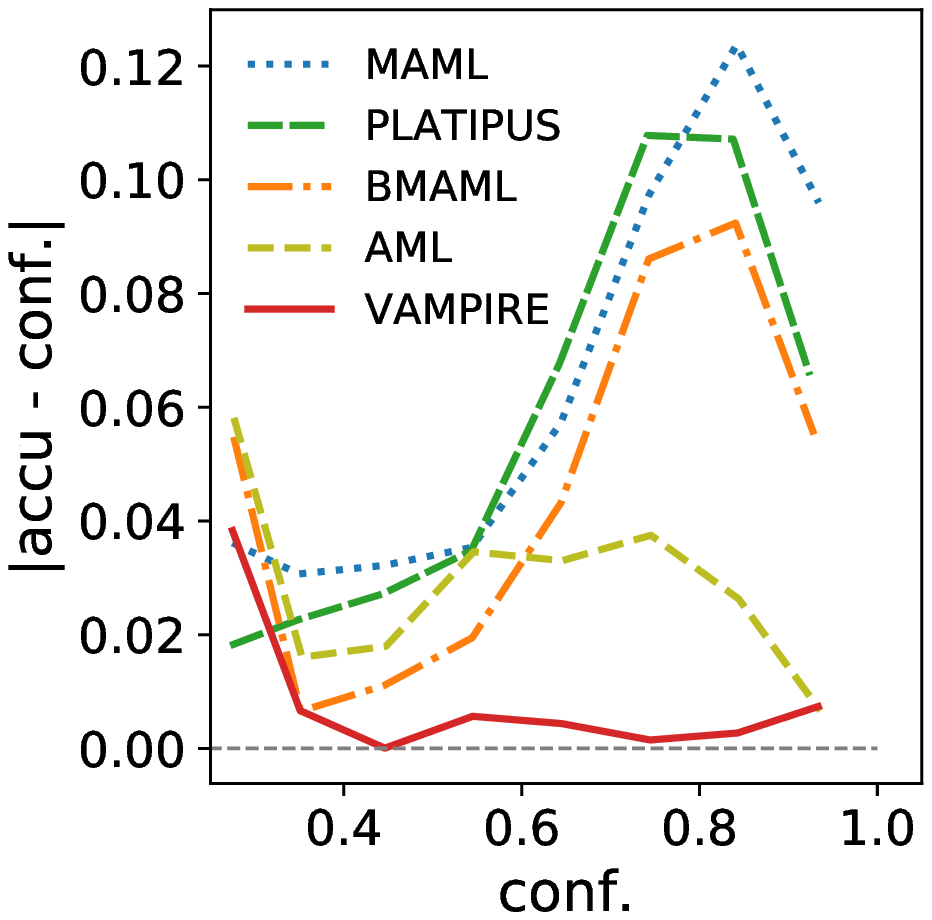}
                \caption{}
                \label{fig:classification_reliability}
            \end{subfigure}
            \begin{subfigure}{0.485\linewidth}
                \centering
                \includegraphics[width = \linewidth]{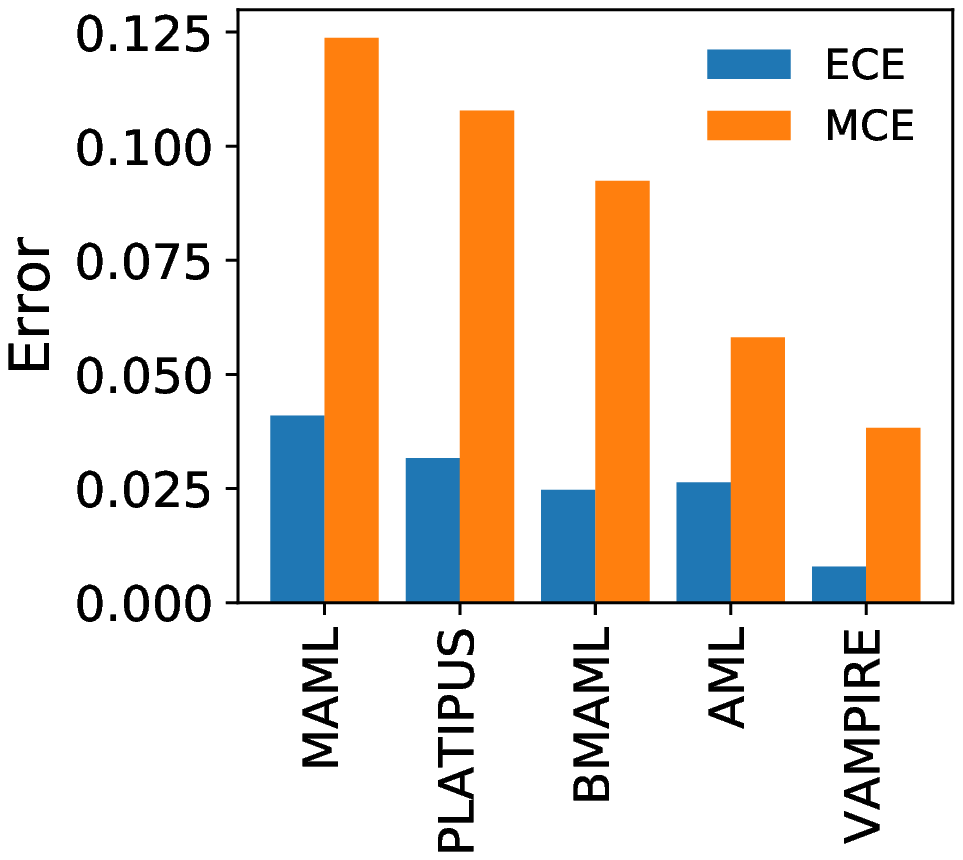}
                \caption{}
                \label{fig:classification_ece_mce}
            \end{subfigure}
            % \vspace{-0.5cm}
            \caption{\protect\subref{fig:classification_reliability} Uncertainty evaluation between different meta-learning methods using reliability diagrams, and \subref{fig:classification_ece_mce} expected calibration error (ECE) and maximum calibration error (MCE), in which the evaluation is carried out on 5-way 1-shot setting for \({20 \choose 5} = 15504\) unseen tasks sampled from mini-ImageNet dataset.}
            \label{fig:reliability_diagram}
        \end{figure}

        To evaluate the predictive uncertainty of the models, we show in \figureautorefname{~\ref{fig:classification_reliability}} the reliability diagrams~\cite{guo2017oncalibration} averaged over many unseen tasks to compare different meta-learning methods. A perfectly calibrated model will have its values overlapped with the identity function \(y=x\), indicating that the probability associated with the label prediction is the same as the true probability. To have a fair comparison, we train all the methods of interest under the same configuration, e.g. network architecture, number of gradient updates, while keeping all method-specific hyper-parameters the same as the reported values. Due to the constrain of GPU memory, BMAML is trained with only 8 particles, while PLATIPUS, Amortimised Meta-learner and VAMPIRE are trained with 10 Monte Carlo samples. According to the reliability graphs, the model trained with VAMPIRE shows a much better calibration than the ones trained with the other methods used in the comparison. To further evaluate, we compute the expected calibration error (ECE) and maximum calibration error (MCE)~\cite{guo2017oncalibration} of each models trained with these methods. Intuitively, ECE is the weighted average error, while MCE is the largest error. The results plotted in \figureautorefname{~\ref{fig:classification_ece_mce}} show that the model trained with VAMPIRE has smaller ECE and MCE compared to all the state-of-the-art meta-learning methods. The slightly low performance of Amortised Meta-learner might be due to the usage of the whole task-specific dataset, potentially overfitting to the training data. Another factor contributed might be the arguable Dirac-delta hyper-prior used, which can be also the cause for the low prediction accuracy shown in \tableautorefname{~\ref{tab:miniImageNet_results}}.
        
    \section{Conclusion}
    We introduce and formulate a new Bayesian algorithm used for few-shot meta-learning. The proposed algorithm, VAMPIRE, employs variational inference to optimise a well-defined cost function to learn a distribution of model parameters. The uncertainty, in the form of the learnt distribution, can introduce more variability into the decision made by the model, resulting in well-calibrated and highly-accurate prediction. The algorithm can be combined with different models that are trainable with gradient-based optimisation, and is applicable in regression and classification. We demonstrate that the algorithm can make reasonable predictions about unseen data in a multi-modal 5-shot learning regression problem, and achieve state-of-the-art calibration and classification results with only 1 or 5 training examples per class on public image data sets.
    {\small\printbibliography}
    \newpage
    \onecolumn
\begin{appendices}
    \section*{SUPPLEMENTARY MATERIAL}
    \section{Regression experiments}
    \label{apdx:regression}
        \subsection{Training configuration}
            As mentioned in \sectionautorefname{~\ref{sec:regression}}, the experiment is carried out on a multi-modal structured data, where a half of tasks are generated from sinusoidal functions, while the other half of tasks are from linear functions. The sinusoidal functions are in the form of \(A\sin(x + \varphi)\), where the amplitude \(A\) and the phase \(\varphi\) are uniformly sampled from [0.1, 5] and [0, \(\pi\)], respectively. The linear functions are in the form of \(ax + b\), where the slope \(a\) and the intercept \(b\) are sampled from the uniform distribution on [-3, 3]. The input \(x\) is uniformly sampled from [-5, 5]. In addition, a Gaussian noise with zero-ed mean and a standard deviation of 0.3 is added to the output.
            
            The model used in this experiment is a 3-hidden fully connected neural network with 100 hidden units per each hidden layer. Output from each layer is activated by ReLU without batch normalisation. The optimisation for the objective function in~\eqref{eq:log_likelihood} is carried out by Adam. Note that for regression, there is we do not place any weighting factor for the KL divergence term of VFE. Please refer to \tableautorefname{~\ref{tab:regression_hyperparameters}} for the details of hyperparameters used.
            
            \begin{table}[ht]
                \centering
                \begin{tabular}{l c c}
                    \toprule
                    \bfseries Hyperparameters & \bfseries Notation & \bfseries Value \\
                    \midrule
                    Learning rate for variational parameters & \(\alpha\) & 0.001 \\
                    Number of gradient updates for variational parameters & & 5 \\
                    Number of Monte Carlo samples for variational parameters & \(L_{t}\) & 128 \\
                    Number of tasks before updating meta-parameters & \(T\) & 10 \\
                    Learning rate for meta-parameters & \(\gamma\) & 0.001 \\
                    Number of Monte Carlo samples for meta-parameters & \(L_{v}\) & 128 \\
                    \bottomrule
                \end{tabular}
                \caption{Hyperparameters used in the regression experiments on multi-modal structured data.}
                \label{tab:regression_hyperparameters}
            \end{table}
        \subsection{Additional results}
            \begin{figure}[ht]
                \centering
                \begin{subfigure}[t]{0.24\linewidth}
                    \centering
                    \includegraphics[width = \linewidth]{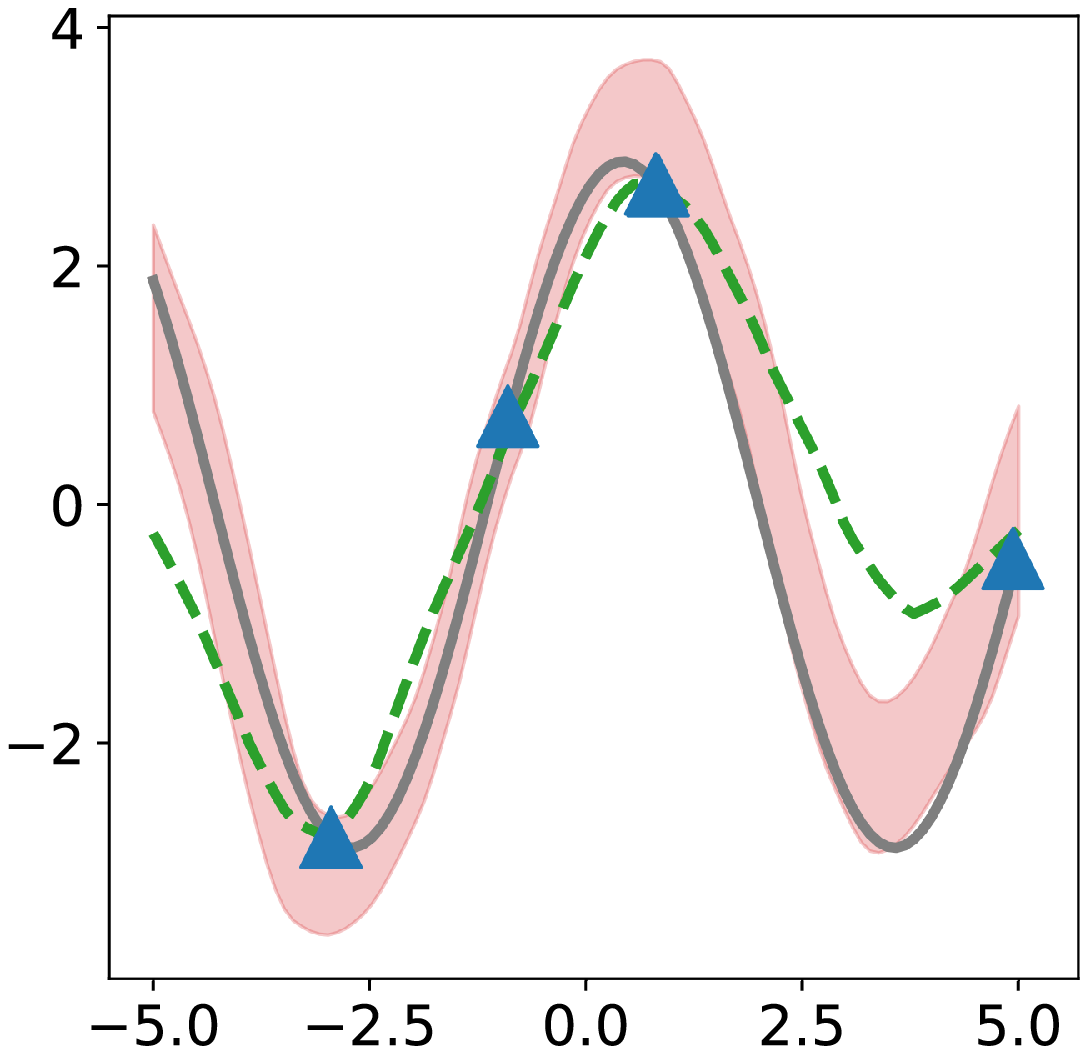}
                \end{subfigure}
                \begin{subfigure}[t]{0.24\linewidth}
                    \centering
                    \includegraphics[width = \linewidth]{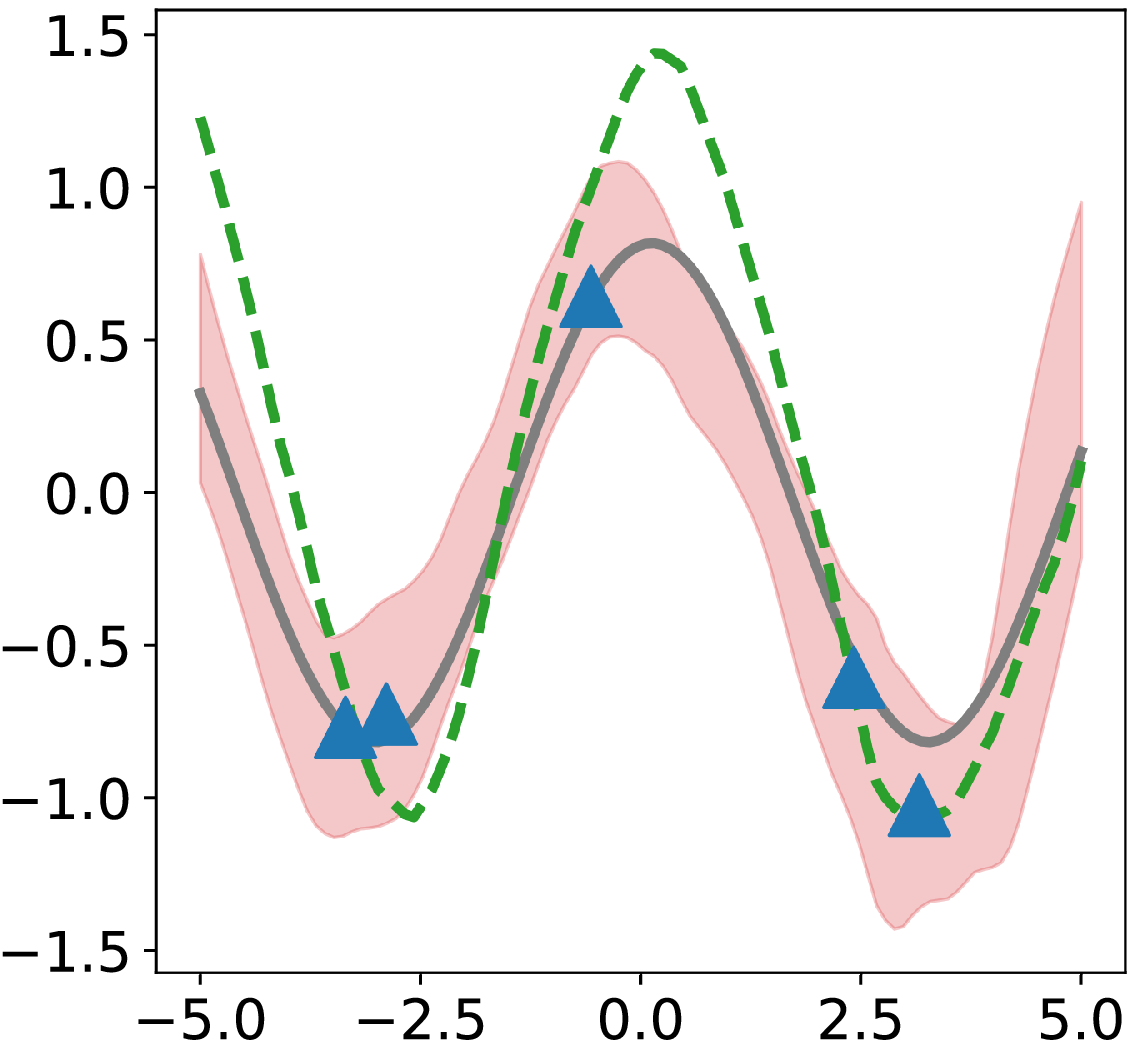}
                \end{subfigure}
                \begin{subfigure}[t]{0.24\linewidth}
                    \centering
                    \includegraphics[width = \linewidth]{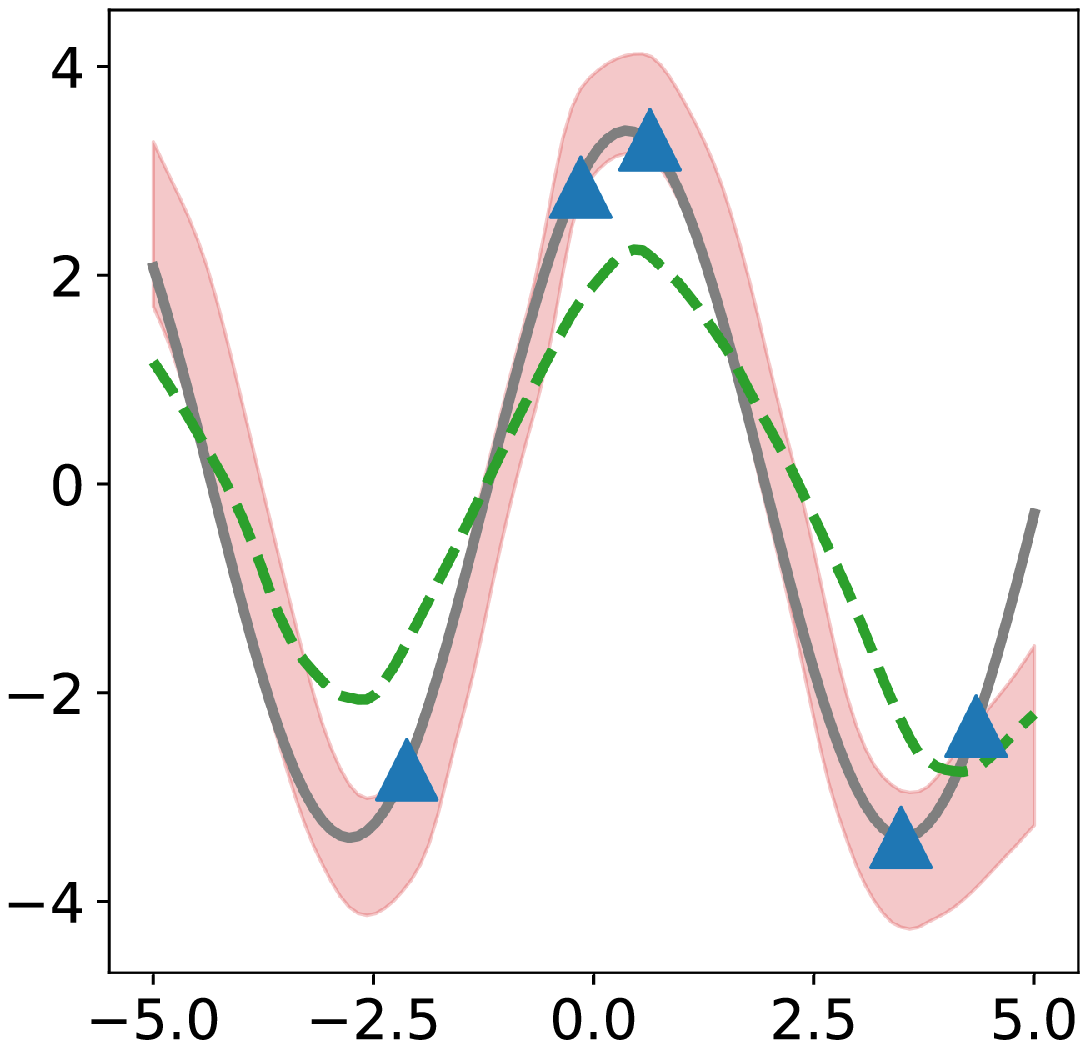}
                \end{subfigure}
                \begin{subfigure}[t]{0.24\linewidth}
                    \centering
                    \includegraphics[width = \linewidth]{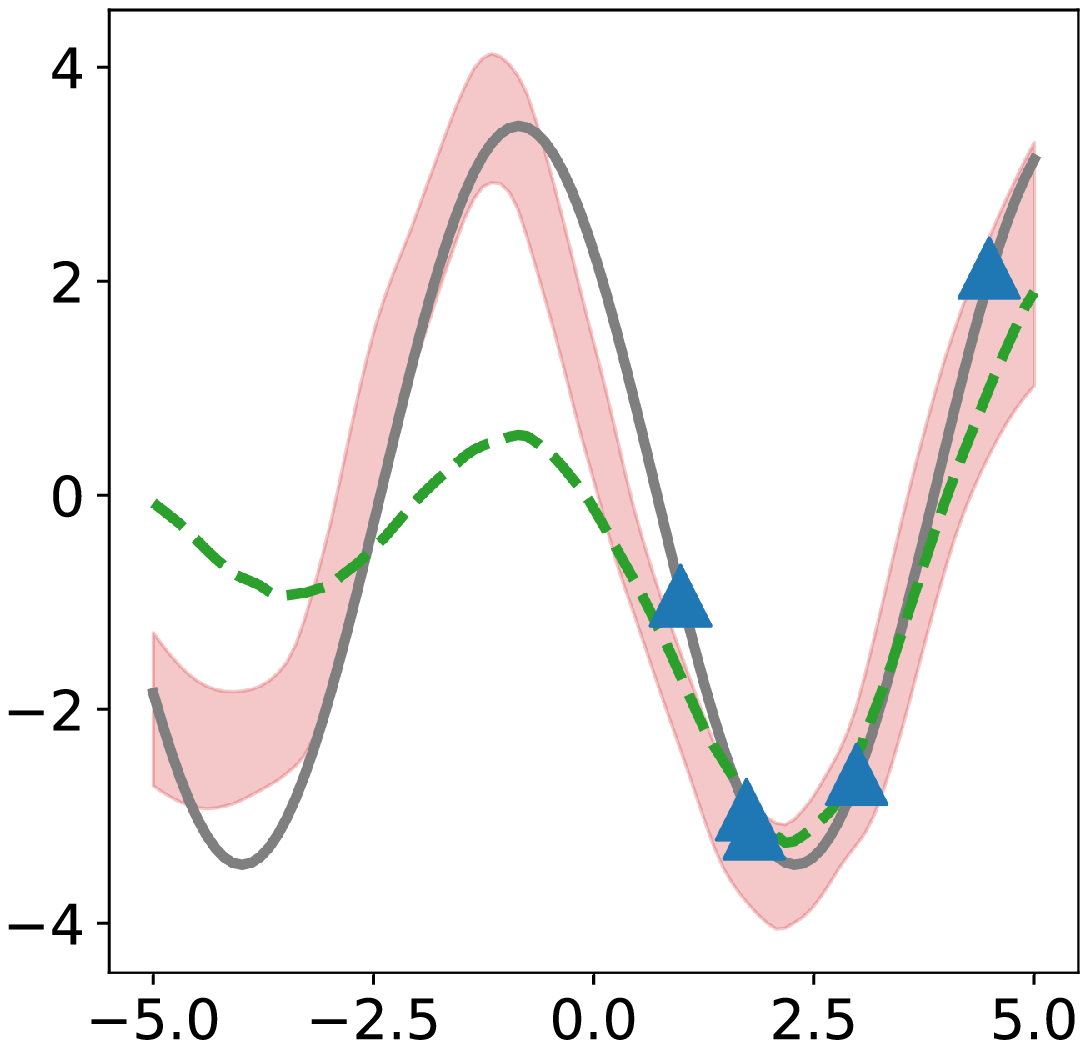}
                \end{subfigure}
                \hfill\\
                \begin{subfigure}[t]{0.24\linewidth}
                    \centering
                    \includegraphics[width = \linewidth]{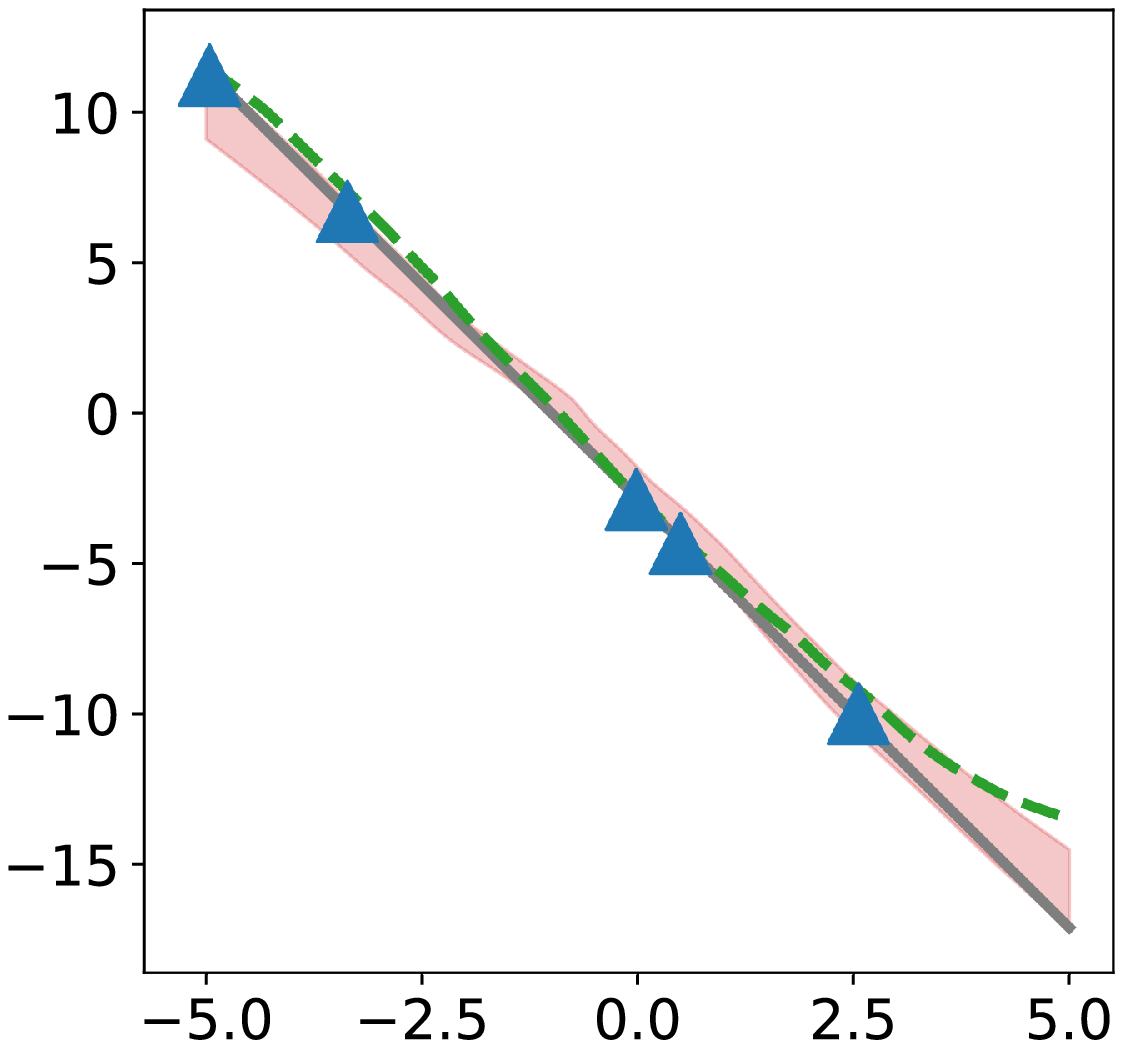}
                \end{subfigure}
                \begin{subfigure}[t]{0.24\linewidth}
                    \centering
                    \includegraphics[width = \linewidth]{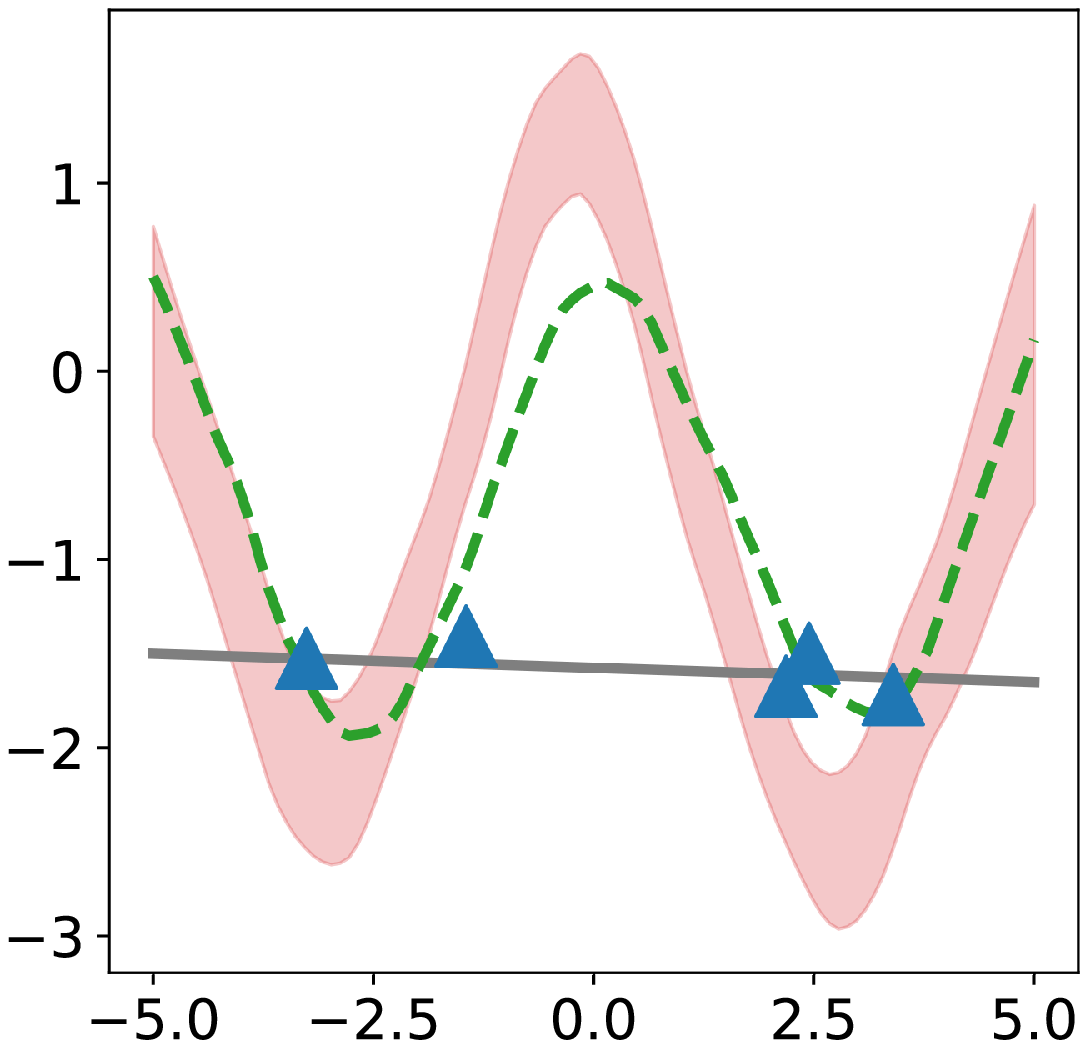}
                \end{subfigure}
                \begin{subfigure}[t]{0.24\linewidth}
                    \centering
                    \includegraphics[width = \linewidth]{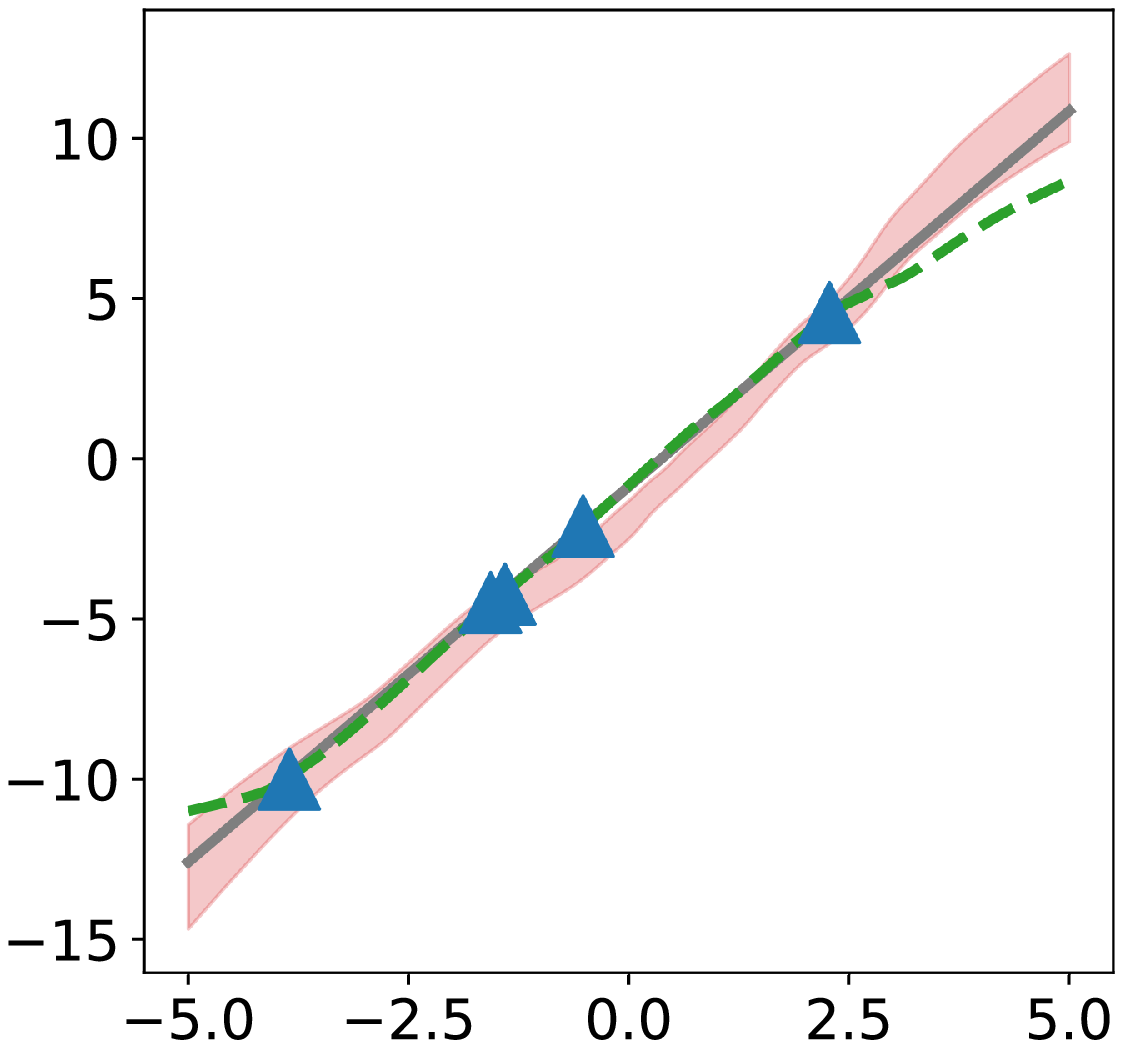}
                \end{subfigure}
                \begin{subfigure}[t]{0.24\linewidth}
                    \centering
                    \includegraphics[width = \linewidth]{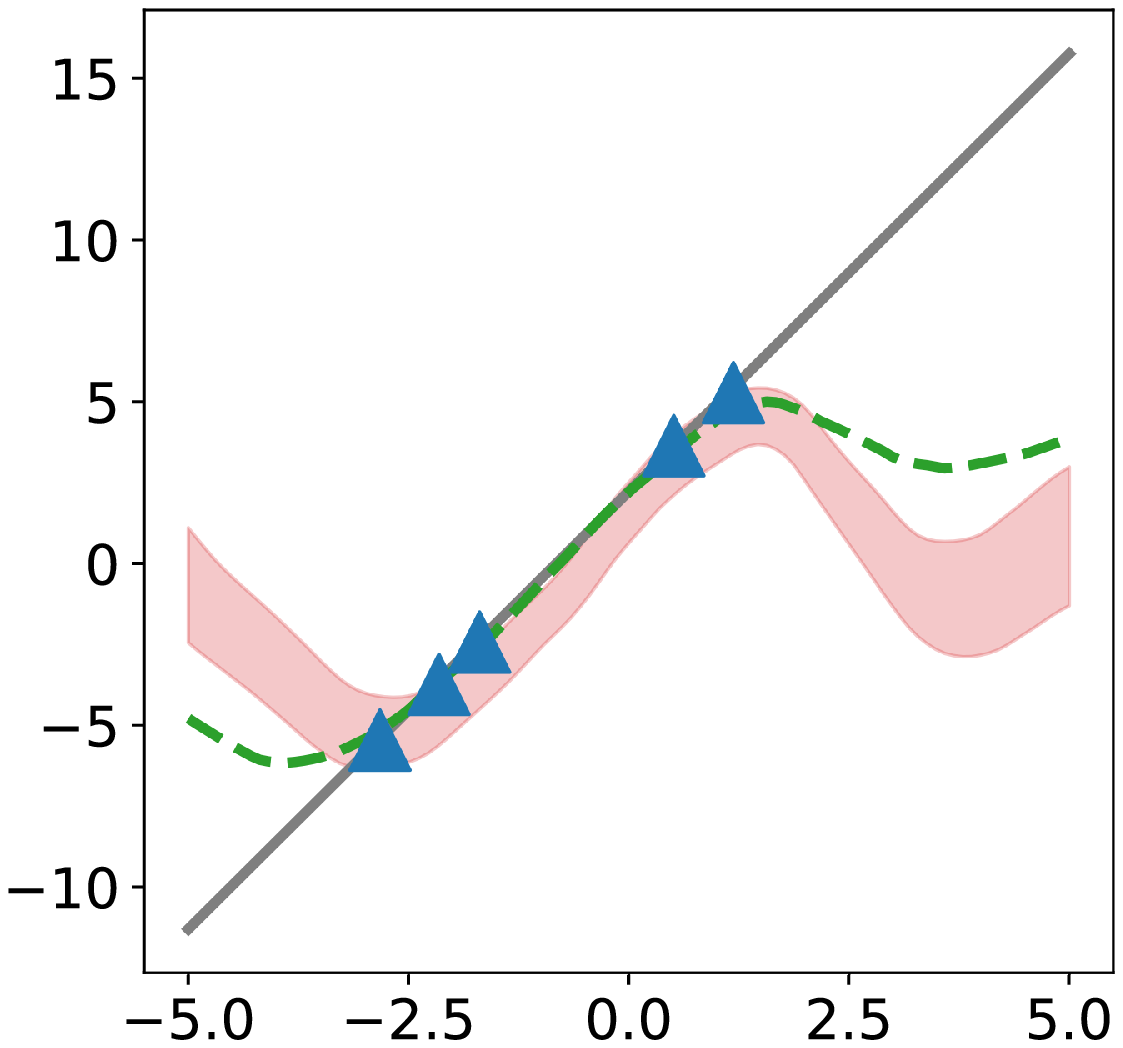}
                \end{subfigure}
                \hfill\\
                \includegraphics[height= 1.25em]{img/legend_mixed_maml.eps}
                \caption{Additional qualitative results with tasks generated from either sinusoidal or linear function. The shaded area is the prediction made by VAMPIRE \(\pm\) 1\(\times\) standard deviation.}
                \label{fig:regression_sine_line_additional}
            \end{figure}
            In addition to the results in \figureautorefname~\ref{fig:regression_sine_line}, we also provide more qualitative visualisation from the multi-modal task distribution in \figureautorefname~\ref{fig:regression_sine_line_additional}.
            
            We also implement many Bayesian meta-learning methods, such as PLATIPUS, BMAML and Amortised Meta-learner, to compare with VAMPIRE using reliability diagram. We train all the methods of interest in the same setting used for VAMPIRE to obtain a fair comparison. The mean-squared error (MSE) of each method after training can be referred to \tableautorefname~\ref{tab:regression_mse}. Please note that for probabilistic methods, MSE is the average value across many Monte Carlo samples or particles sampled from the posterior distribution of model parameters.
            
            \begin{table}[ht]
                \centering
                \begin{tabular}{l c}
                    \toprule
                    \bfseries Method & \bfseries MSE \\
                    \midrule
                    MAML & 1.96 \\
                    PLATIPUS & 1.86 \\
                    BMAML & 1.12 \\
                    Amortised Meta-learner & 2.32\\
                    VAMPIRE & 2.24 \\
                    \bottomrule
                \end{tabular}
                \caption{Mean squared error of many meta-learning methods after being trained in the same setting are tested on 1000 tasks.}
                \label{tab:regression_mse}
            \end{table}
            
        % \subsection{Model calibration for regression - ECE and MCE}
        %     We provide the numeric form for the ECE and MCE shown in \figureautorefname~\ref{fig:regression_ece_mce}, including 95\% confidence interval in \tableautorefname~\ref{tab:regression_ece_mce}.
            
        %     \begin{table}[ht]
        %         \centering
        %         \begin{tabular}{l c c}
        %             \toprule
        %             \bfseries Method & \bfseries ECE & \bfseries MCE \\
        %             \midrule
        %             PLATIPUS &  & \\
        %             BMAML & & \\
        %             Amortised Meta-learner & & \\
        %             VAMPIRE & & \\
        %             \bottomrule
        %         \end{tabular}
        %         \caption{Caption}
        %         \label{tab:regression_ece_mce}
        %     \end{table}
        
    \section{Classification experiments}
    \label{apdx:classification}
        This section describes the detailed setup to train and validate the few-shot learning on Omniglot and mini-ImageNet presented in Sec.~\ref{sec:classification}.
        Following the notation used in Sec.~\ref{sec:few_shot_problem}, each task or episode \(i\) has \(N\) classes, where the support set \(\mathcal{Y}_{i}^{(t)}\) has $k$ samples per class, and the query set \(\mathcal{Y}_{i}^{(v)}\) has 15 samples per class. This is to be consistent with the previous works in the literature \cite{ravi2017optimization, finn2017model}. The training is carried out by using Adam to minimise the cross-entropy loss of the softmax output. The learning rate of the meta-parameters \(\theta\) is set to be \(\gamma=10^{-3}\) across all trainings, and decayed by a factor of 0.99 after every 10,000 tasks. Other hyperparameters used are specified in \tableautorefname{~\ref{tab:hyperparameter}}. We select the number of ensemble models \(L_{t}\) and \(L_{v}\) to fit into the memory of one Nvidia 1080 Ti GPU. Higher values of \(L_{t}\) and \(L_{v}\) are desirable to achieve a better Monte Carlo approximation.
        
        \begin{table}[ht]
            \begin{center}
                \begin{small}
                    \begin{sc}
                        \begin{tabular}{l c c c c}
                            \toprule
                            \multirow{2}{*}{\bfseries Description} & \multirow{2}{*}{\bfseries Notation} & \multicolumn{2}{c}{\bfseries Omniglot} & \bfseries mini-ImageNet \\
                            & & 5-way & 20-way & 5-way \\
                            \midrule
                            Number tasks per meta-update & $T$ & 32 & 16 & 2 \\
                            Number of ensemble models (train) & $L_{t}$ (train) & 1 & 1 & 10 \\
                            Number of ensemble models (train) & $L_{v}$ (train) & 1 & 1 & 10 \\
                            Number of ensemble models (test) & $L_{t}$ (test) & 10 & 10 & 10 \\
                            Number of ensemble models (test) & $L_{v}$ (test) & 10 & 10 & 10 \\
                            Learning rate for \(\mathbf{w}_{i}\) & $\alpha$ & 0.1 & 0.1 & 0.01 \\
                            Learning rate for \(\theta\) & $\gamma$ & \(10^{-3}\) & \(10^{-3}\) & \(10^{-3}\) \\
                            Number of inner gradient updates & & 5 & 5 & 5 \\
                            % \midrule
                            % L2 regularisation for $\bm{\mu}_{\theta}$ & & \(10^{-5}\) & \(10^{-5}\) & \(10^{-5}\) \\
                            % L2 regularisation for $\bm{\rho}_{\theta}$ & & 0 & 0 & \(10^{-6}\) \\
                            \bottomrule
                        \end{tabular}
                    \end{sc}
                \end{small}
            \end{center}
            \caption{Hyperparameters used in the few-shot classification presented in Sec.~\ref{sec:experiments}.}
            \label{tab:hyperparameter}
        \end{table}
        
        For the experiments using extracted features \cite{rusu2019meta} presented in \tableautorefname{~\ref{tab:miniImageNet_results_nonstandard_nets}} for mini-ImageNet, and the bottom part of \tableautorefname{~\ref{tab:miniImageNet_results}} for tiered-ImageNet, we used a 2-hidden fully connected layer with 128 and 32 hidden units. The learning rate \(\alpha\) is set as 0.01 and 5 gradient updates were carried out. The learning rate for meta-parameters was \(\gamma = 0.001\).
        
        Both the experiments for classification re-weight the KL divergence term of VFE by a factor of 0.1.
        
        \begin{table}[ht]
            \begin{center}
                \begin{small}
                    \begin{sc}
                        \begin{tabular}{l c c}
                            \toprule
                            \multirow{2}{*}{} & \multicolumn{2}{c}{\textbf{Mini-ImageNet}~\cite{ravi2017optimization}} \\
                             & 1-shot & 5-shot\\
                            \midrule
                            \multicolumn{3}{l}{\textbf{Non-standard CNN}} \\
                            \midrule
                            Relation nets \cite{Sung_2018_CVPR} & 50.44 $\pm$ 0.82 & 65.32 $\pm$ 0.70 \\
                            VERSA \cite{gordon2018metalearning} & 53.40 $\pm$ 1.82 & 67.37 $\pm$ 0.86 \\
                            % Meta Nets \cite{munkhdalai2017meta} & 49.21 $\pm$ 0.96 & \_ \\
                            SNAIL \cite{mishra2018simple} & 55.71 $\pm$ 0.99 & 68.88 $\pm$ 0.92 \\
                            adaResNet \cite{munkhdalai2018rapid} & 56.88 $\pm$ 0.62 & 71.94 $\pm$ 0.57 \\
                            TADAM \cite{oreshkin2018tadam} & 58.5 $\pm$ 0.30 & 76.7 $\pm$ 0.30 \\
                            LEO \cite{rusu2019meta} & 61.76 $\pm$ 0.08 & 77.59 $\pm$ 0.12 \\
                            MetaOptNet \cite{lee2019meta} & \bfseries 64.09 \(\pm\) 0.62 & \bfseries 80.00 \(\pm\) 0.45 \\
                            \rowcolor{gray!30} \textbf{VAMPIRE} & 62.16 \(\pm\) 0.24 & 76.72 \(\pm\) 0.37 \\
                            \bottomrule
                        \end{tabular}
                    \end{sc}
                \end{small}
            \end{center}
            % \vspace{-1em}
            \caption{Accuracy for 5-way classification on mini-ImageNet tasks (in percentage) of many methods which uses extra parameters, deeper network architectures or different training settings.}
            \label{tab:miniImageNet_results_nonstandard_nets}
        \end{table}
    
    \subsection{Model calibration for classification - ECE and MCE}
        We provide the results of model calibration, in particular, ECE and MCE in the numeric form. We also include the 95\% confidence interval in \tableautorefname~\ref{tab:classification_ece_mce}, although they are extremely small due to the large number of unseen tasks.
        
        \begin{table}[ht]
            \centering
            \begin{tabular}{l c c}
                \toprule
                \bfseries Method & \bfseries ECE & \bfseries MCE \\
                \midrule
                MAML & \(0.0410 \pm 0.005\) & 0.124 \\
                PLATIPUS & \(0.032 \pm 0.005\) & 0.108 \\
                BMAML & \(0.025 \pm 0.006\) & 0.092 \\
                Amortised Meta-learner & \(0.026 \pm 0.003\) & 0.058 \\
                VAMPIRE & \(0.008 \pm 0.002\) & 0.038 \\
                \bottomrule
            \end{tabular}
            \caption{Results of ECE and MCE of several meta-learning methods that are tested in 5-way 1-shot setting over 15504 unseen tasks sampled from mini-ImageNet dataset.}
            \label{tab:classification_ece_mce}
        \end{table}

    \section{Pseudo-code for evaluation}
    \label{sec:algorithm_test}
        \begin{algorithm}[htb]
        	\caption{VAMPIRE testing}
        	\label{alg:vampire_test}
        	\begin{algorithmic}[1]
        		\REQUIRE a new task \(\mathcal{T}_{T+1}\),  $\theta$, $L_{t}, L_{v}$, $\alpha$ and $\beta$
        		\STATE $\lambda_{T+1} \gets \bm{\theta}$
        		\STATE sample $\hat{\mathbf{w}}_{T+1}^{(l)} \sim q(\mathbf{w}_{T+1} \vert \lambda_{T+1})$, where $l_{t}=1:L_{t}$
        		\STATE update: $\lambda_{i} \gets \lambda_{i} - \frac{\alpha}{L_{t}} \nabla_{\lambda_{i}} \mathcal{L}_{i}^{(t)} \vert_{\mathcal{Y}^{(t)}_{T+1}}$
        		\STATE draw $L_v$ ensemble model parameters $\hat{\mathbf{w}}_{i}^{(l_v)} \sim q(\mathbf{w}_{i}; \lambda_{i})$
        		\STATE compute prediction $\mathcal{\hat{Y}}_{i}^{(v)}$ using $L_{v}$ ensemble models.
        	\end{algorithmic}
        \end{algorithm}
\end{appendices}
\end{document}